\documentclass[dvipsnames]{article}
\usepackage{arxiv}

\usepackage[utf8]{inputenc}
\usepackage{amsmath}
\usepackage{amsthm}
\usepackage{graphicx}
\usepackage[toc,page]{appendix}
\usepackage{amssymb}
\usepackage{url}
\usepackage{enumitem}
\usepackage{dsfont}
\usepackage{breakcites}
\usepackage{xcolor}
\usepackage[colorlinks=true,linkcolor=ForestGreen,breaklinks]{hyperref}
\usepackage{bm}
\usepackage{multirow}

\usepackage{xcolor}
\usepackage{booktabs}       
\usepackage{amsfonts}       
\usepackage{nicefrac}       
\usepackage{microtype}      
\usepackage{float}
\usepackage{subfig}
\usepackage[percent]{overpic}
\usepackage{hhline}
\usepackage{times} 
\usepackage{url}
\usepackage{tikz}

\newcommand{\ie}{i.e.}

\newtheorem{definition}{Definition}
\newtheorem{theorem}{Theorem}

\newtheorem{corollary}{Corollary}
\newtheorem{remark}{Remark}

\newcommand{\Gu}{\texttt{G1}}
\newcommand{\Gd}{\texttt{G2}}
\newcommand{\Gt}{\texttt{G3}}
\newcommand{\Gq}{\texttt{G4}}
\newcommand{\Gc}{\texttt{G5}}

\newcommand{\argmax}[1]{
    \underset{#1}{\mathrm{argmax}}\
}

\DeclareMathOperator{\tr}{trace}
\DeclareMathOperator{\diag}{diag}

\newcommand{\argmin}[1]{
    \underset{#1}{\mathrm{argmin}}\
}

\allowdisplaybreaks

\title{All of the Fairness for Edge Prediction with Optimal Transport}

\author{Charlotte Laclau, Ievgen Redko, Manvi Choudhary, Christine Largeron \\
  Univ Lyon, UJM-Saint-Etienne, CNRS, Institut d'Optique Graduate School\\
  Laboratoire Hubert Curien UMR 5516, F-42023, Saint-Etienne, France\\
  \url{name.surname@univ-st-etienne.fr}}

\begin{document}
\maketitle

\begin{abstract}
Machine learning and data mining algorithms have been increasingly used recently to support decision-making systems in many areas of high societal importance such as healthcare, education, or security. While being very efficient in their predictive abilities, the deployed algorithms sometimes tend to learn an inductive model with a discriminative bias due to the presence of this latter in the learning sample. This problem gave rise to a new field of algorithmic fairness where the goal is to correct the discriminative bias introduced by a certain attribute in order to decorrelate it from the model's output. In this paper, we study the problem of fairness for the task of edge prediction in graphs, a largely underinvestigated scenario compared to a more popular setting of fair classification. To this end, we formulate the problem of fair edge prediction, analyze it theoretically, and propose an embedding-agnostic repairing procedure for the adjacency matrix of an arbitrary graph with a trade-off between the group and individual fairness. We experimentally show the versatility of our approach and its capacity to provide explicit control over different notions of fairness and prediction accuracy. 
\end{abstract}

\section{Introduction}
We live in a world where an increasing number of decisions, with major societal consequences, are made or at least supported by algorithms that diligently learn the patterns from a training sample and gain their discriminating ability by identifying the key attributes correlated with the desired output. These attributes, however, can represent sensitive information that, in its turn, can lead to a significant bias in model's predictions when deployed on a previously unseen sample. For instance, when building a recommendation system supporting a recruitment company in finding a potential candidate suitable for their clients' needs, one would expect its recommendations to be independent from the gender or the ethnicity of the considered individuals. In practice, however, the training sample used to learn the model may have been collected in a biased manner with an unequal number of successive outcomes between the genders and/or ethnic groups. The recommendations of the learned model in this case will tend to follow the learned pattern thus reinforcing the already existent bias. Research works aiming at identifying and correcting such inductive bias form the core of the algorithmic fairness field, a scientific area that is constantly gaining more and more attention from the machine learning and data mining communities nowadays. 

Algorithmic fairness methods are traditionally divided into one of the three following categories: (i) pre-processing methods that repair the original data to remove the bias, ii) methods that integrate fairness constraints or penalties in a given learning algorithm and iii) post-processing methods that debias directly the model's output. First family of methods can be further divided into two subfamilies where the first one corrects the input raw data to ensure that the inference of the sensitive attribute is impossible, regardless of the learning algorithm (e.g. classifier) used downstream \cite{Feldman2015,Calmon2017,johndrow2019}, while the other learns a new representation that is forced to be independent from the sensitive attribute \cite{Zemel2013,Edwards2016,Louizos2016,Madras18}. These methods present the most generic solution to the considered problem as they allow to use any available algorithm on the repaired data and ensure that the generalization performance on this latter would be comparable to that obtained on the original data \cite{Gordaliza19}. 
The methods belonging to the second category \cite{zafar2017a,zafar2017b,Corbett2017,agarwal2018a,Donini2018} are specific to the learning algorithm and thus the modification of this latter due to, for instance, a performance drop on another data set requires modifying the whole optimization procedure. Finally, the last category \cite{Hardt2016,Kusner2017,Jiang2019,Chiappa19,ZehlikeHW20} of methods has a virtue of debiasing directly the outputs of a learning algorithm, but similarly to the methods that impose fairness constraints while learning require the post-processing to be performed for each prediction. 

Most of the works mentioned above address the problem of algorithmic fairness in the context of supervised classification and completely ignore learning from relational data given in form of structured objects or graphs, and the tasks associated to it. Such data, however, is ubiquitous in areas dealing with complex systems, especially in the social sciences where the relationships and interactions between people are studied. Several mining tasks can be defined for such data, such as edge prediction, node classification or community detection, to name a few.

\paragraph{Contributions}
In this paper, we propose a first theoretically sound embedding-agnostic method for group and individually fair edge prediction. This is done through the following contributions:
\begin{enumerate}[ labelwidth=!, labelindent=0pt]
    \item We analyze the group fair edge prediction task theoretically and show that one can efficiently repair the adjacency matrix of a graph by aligning the joint distributions of nodes appearing in different sensitive groups. 
    \item We derive an optimal transport (OT)-based algorithm from our analysis, add individual fairness constraints to it and implement it for binary and multi-class settings. 
    The proposed algorithm outputs a repaired adjacency matrix by adding edges that obfuscate the dependence on the sensitive attribute. The repaired adjacency matrix can be used as input of any node embedding technique thus making it embedding agnostic. 

    \item We evaluate the efficiency of our approach through extensive experiments on several synthetic and real-world data sets and show that it provides an explicit control on the trade-off between the two notions of fairness and prediction accuracy.
\end{enumerate}

\paragraph{Organisation}
The rest of this paper is organized as follows. 
We provide a theoretical analysis of group fair edge prediction in Section \ref{sec:contribution}. In Section \ref{sec:algo}, we present a group and individually fair repair scheme for adjacency matrix of a graph. In Section \ref{sec:results}, we evaluate our approach on synthetic and real-world networks and show the impact of the proposed repairing scheme both on the capacity of predicting the sensitive attribute from embeddings learned with repaired data and on the performance of edge prediction. Last section concludes the paper and gives a couple of hints for possible future research. 

\section{Fair edge prediction}\label{sec:contribution}
In this section, we formulate the problem of fair edge prediction in graphs and give a definition of several key concepts related to it such as statistical parity, disparate impact and balanced error rate. We further analyze this problem and derive a theoretically sound approach allowing to solve it. 
\subsection{Problem setup}
Let $\mathbb{V}$ denote an abstract vertex space, and let $\mathcal{V} = \{v_1, \dots, v_{N}\} \in \mathbb{V}^N$ be a set of $N$ vertices drawn independently and identically (iid) from an arbitrary  distribution over $\mathbb{V}$. Let $r: \mathbb{V}\times \mathbb{V} \rightarrow \{0,1\}$ be a (symmetric) true edge prediction function that outputs $1$ if there is an edge between two nodes and $0$ otherwise. In the finite case, we further consider a graph $\mathcal{G}=(\mathcal{V}, \mathcal{E})$
where $\mathcal{E} \subseteq \mathbb{V}\times \mathbb{V}$ is labeled according to $r$ such that a tuple $\{(v_i, v_j, r(v_i,v_j)\}$ defines an undirected edge $(i,j) \in \mathcal{E}$. 
Furthermore, we assume that all nodes have one categorical sensitive attribute $S:\mathbb{V} \rightarrow \mathbb{S}$ where, for simplicity, we assume that $\mathbb{S}$ is a set $\{0,1\}$. In the context of fair edge prediction, this variable defines a potential source of \textit{bias} in the graph where $S = 0$ stands for the minority (unfavored) class, while $S = 1$ stands for the default (favored) class. In what follows, we are interested in the edge prediction task where the goal is to find a function $h:\mathbb{V}\times \mathbb{V} \rightarrow \{0,1\}$ such that $h$ is as close as possible to $r$. 

We define the notion of \textit{statistical parity} of $h$ for this scenario as the equality between the probability of $h$ for predicting the same value, say 1, for both nodes belonging to the same and different classes. More formally, this definition is given below.
\begin{definition} Given a graph $\mathcal{G}=(\mathcal{V}, \mathcal{E})$ and a function $h:\mathbb{V} \times \mathbb{V} \rightarrow \{0,1\}$, we define the statistical parity for an edge predictor $h$ on $S$ with respect to (w.r.t) $\mathbb{V}$ as: 
    $$\mathbb{P}(h(V,V')=1|S \neq S') = \mathbb{P}(h(V,V')=1|S = S')$$
    or equivalently
    $$\mathbb{P}(h(V,V')=1|S \oplus S' = 1) = \mathbb{P}(h(V,V')=1|S \oplus S' = 0),$$
	\label{def:stat_parity}
	where $\oplus$ stands for XOR operation and the probability is taken over random variables $((V,S),(V',S')) \sim \mathcal{D}\times \mathcal{D}$ with $\mathcal{D}$ denoting the joint distribution over $\mathbb{V}\times \mathbb{S}$.
\end{definition}
This definition states that the probability for $h$ to predict an edge between two nodes $v$ and $v'$ is the same whether $v$ and $v'$ belong to the same ($S = 0,\ S' = 0$ or $S = 1,\ S' = 1$) or to the different ($S = 1,\ S' = 0$ or $S = 0,\ S' = 1$) classes. One may note that it is different from the definition considered in (node) classification task with $\mathbb{V} \subset \mathbb{R}^d,\ h:\mathbb{V} \rightarrow\{0,1\}$ as this latter trivially reduces to a usual fair classification problem studied extensively in the literature. In our case, however, we have to deal with tuples of variables and implicitly attribute a sensitive variable defined by $S \oplus S'$ to each pair of nodes or, equivalently, to an edge. On a higher level, this transposes the initial problem into repairing adjacency matrices contrary to repairing the feature representation of nodes as done in the case of node classification.
Bearing in mind the equivalent XOR representation, we further denote these two events by 
\begin{align}
    \mathbb{P}_1(h) &= \mathbb{P}(h(V,V')=1|S \neq S'),\notag\\
    \mathbb{P}_0(h) &= \mathbb{P}(h(V,V')=1|S = S'). 
    \label{eq:p1_p0}
\end{align}
Using these notations, we define two other important fairness measures, notably \textit{disparate impact} ($\text{DI}$) and \textit{balanced error rate} ($\text{BER}$). We give their definitions below.
\begin{definition}
Given a graph $\mathcal{G}=(\mathcal{V}, \mathcal{E})$ and a function $h:\mathbb{V} \times \mathbb{V} \rightarrow \{0,1\}$, let $\mathbb{P}_0(h)$ and $\mathbb{P}_1(h)$ be defined as in \eqref{eq:p1_p0}. We define disparate impact ($\text{DI}$) and balanced error rate ($\text{BER}$) for an edge predictor $h$ on $S\oplus S'$ w.r.t. $\mathbb{V}$ as: 
\begin{align*}
	\text{DI}(h,\mathbb{V},S\oplus S')&= \frac{\mathbb{P}_1(h)}{\mathbb{P}_0(h)},\\
	\text{BER}(h,\mathbb{V},S\oplus S') &= \frac{\mathbb{P}_1(h)-\mathbb{P}_0(h)+1}{2}.
\end{align*}
\end{definition}
Each of these two measures has its own interpretation. DI is identical to statistical parity and it is equal to 1, when $h$ is perfectly fair. In practice, we are interested in bounding this latter, \ie, $\text{DI}(h,\mathbb{V},S\oplus S') \leq \tau$, indicating that a classifier has a disparate impact at level $\tau \in (0,1]$. As for the BER, it stands for a misclassification error of the sensitive attribute $S$ by $h$ in a setting where $\mathbb{P}(S\oplus S' = 1) = \mathbb{P}(S\oplus S' = 0)$. This latter condition roughly tells us that the probability of drawing a pair of nodes belonging to the same class should be the same as the probability of them belonging to different classes. It is important to note that while higher values of disparate impact and $\tau$ indicate a more fair outcome, the best misclassification error in terms of fairness is equal to $\frac{1}{2}$ as in this case a classifier is not capable of predicting whether the nodes are from the same or different classes.
\begin{remark}
In what follows, it will be also convenient to define both the disparate impact and the balanced error rate w.r.t. $S$ by considering only the conditioning on one variable of the pair $(S,S')$. For this latter case, we write 
$$\text{DI}(h,\mathbb{V},S) = \frac{\mathbb{P}(h(V,V')=1|S =0)}{\mathbb{P}(h(V,V')=1|S =1)}$$
and similarly for $\text{BER}(h,\mathbb{V},S)$.
\end{remark}
We proceed to the analysis of our fairness setting below.

\subsection{Analysis of group fair edge prediction}
Several works \cite{feldman15,Gordaliza19,Jiang2019} provided a theoretical analysis for the fair classification setting where one deals with one random variable $X: \Omega \rightarrow \mathbb{R}^d$ with $\Omega$ being an arbitrary instance space and considers learning a hypothesis function $h: \mathbb{R}^d \rightarrow\{0,1\}$. Below, we provide the analysis for the edge prediction fairness and relate it to the statistical parity of $h$ in predicting the sensitive attributed individually for one of the node's pair. Note that from the algorithmic point of view, working with edges given by pairs of nodes and their associated sensitive attributes $S \otimes S'$ is hard as they do not admit any representation allowing further repair. To this end, our goal would be to simplify the problem in a principled way by considering learning on the joint space of nodes with the sensitive attribute being related to only one node from a pair. In this case, we would be able to use the pair-wise information about the nodes as node representation. 

To proceed, we first make the following assumptions.
\begin{enumerate}[leftmargin = .7cm]
    \item[\textbf{A1}.] The probability of each node belonging to the favoured or unfavoured class is the same, \ie, 
    $$\mathbb{P}(S=0)\! = \!\mathbb{P}(S'=0)\! = \! \mathbb{P}(S=1)\! = \! \mathbb{P}(S'=1)\! = \! \frac{1}{2}.$$ 
    \item[\textbf{A2}.] The probability of predicting an edge given that both nodes are in the same class is higher than that of predicting an edge between the nodes of different classes, \ie, for all $s\in \{0,1\}$
    \begin{align*}
        \mathbb{P}(h(V,V')=1&|S=s,S'=1-s) \leq \mathbb{P}(h(V,V')=1|S=s,S'=s).
    \end{align*}
\end{enumerate}
We state our main result below\footnote{All proofs are provided in the Appendix.}.
\begin{theorem}
Consider a graph $\mathcal{G}=(\mathcal{V}, \mathcal{E})$, sets $S, S' \in \{0,1\}$, an edge prediction function $h:\mathbb{V} \times \mathbb{V} \rightarrow \{0,1\}$ and assume that $\text{DI}(h,\mathbb{V},S) \leq \tau$ for some $\tau \in (0,1]$. Then, with the assumptions A1-A2 the following holds:
\begin{align*}
     \text{DI}(h,\mathbb{V},S\oplus S') \leq \text{DI}(h,\mathbb{V},S) \leq \tau.
\end{align*}
\label{trm:1}
\end{theorem}
Before discussing the implications of this theorem, we briefly note that the assumptions A1-A2 are not restrictive and capture one's intuition about the fair edge prediction setting. Indeed, A1 assumes that each node has an equal probability of belonging or not to the favoured class which is a reasonable assumption to make in practice when sampling vertices at random. In its turn, A2 says that the probability of predicting an edge inside any of the two classes is higher than that of predicting an edge between different classes. This latter is related to assortativity effect and it is rather intuitive as sensitive attributes are often correlated with a certain latent structure of the graph and thus can be seen as a ``community" indicator of each node. 

As for the implications, several remarks are in order here. First, the theorem shows that the disparate impact w.r.t. the sensitive attribute $S$ of \textit{individual} nodes provides an upper bound on the disparate impact of $h$ when considering the compositional sensitive attribute $S\oplus S'$ defined for pairs of nodes. An immediate consequence of this is that repairing a graph in this case can be done by considering only classes of nodes ($S=0$ and $S=1$) rather that classes of edges ($S=S'$ and $S\neq S'$) given by pairs of nodes. Second, the established inequality allows to further provide several results for $\text{BER}(h,\mathbb{V},S\oplus S')$ that suggest an algorithmic solution for an OT-based repairing procedure. We give this result below. 
\begin{corollary}
With the assumption from Theorem \ref{trm:1}, we have
\begin{align*}
    \text{BER}(h,\mathbb{V},S\oplus S') &\leq \frac{1}{2} - \frac{\mathbb{P}_1(h)}{2}\left(\frac{1}{\tau}-1\right),\\
    \min_{h \in \mathcal{H}} \text{BER}(h,\mathbb{V},S\oplus S') &= \frac{1}{2}(1-\frac{1}{2}W_{\bm{1}_{\cdot\neq \cdot}}(\gamma_0,\gamma_1))
\end{align*}
where $W_{\bm{1}_{\cdot\neq \cdot}}$ is the Wasserstein distance between true joint distributions $\gamma_0,\gamma_1$ over $\mathbb{V}\times \mathbb{V}$ given $S=0$ and $S=1$, respectively equipped with the Hamming cost function.
\label{cor:1}
\end{corollary}
This corollary provides an upper bound on $\text{BER}(h,\mathbb{V},S\oplus S')$ in terms of $\text{DI}(h,\mathbb{V},S)$ and thus allows to control the former by maximizing the latter. Furthermore, the second part of the statement tells us that the balanced error rate of the best edge predictor depends on the divergence between the joint distributions over the nodes given that the sensitive attribute is equal to one of its possible values. This implication allows us to come up with an algorithmic implementation of the repair procedure based on aligning these joint conditional distributions with an OT coupling acting as a mapping. We use this idea as the backbone for our approach and further explore how one can constrain this mapping to ensure individual fairness too.

\section{Algorithmic implementation}\label{sec:algo}
We now present our algorithm, its multiclass extension and its positioning w.r.t. other related works. 
\subsection{Group Graph Fairness with OT} 
The algorithmic idea behind minimizing $W_{\bm{1}_{\cdot\neq \cdot}}(\gamma_0,\gamma_1)$ from Corollary \ref{cor:1} is to find an optimal transportation plan between $\gamma_0$ and $\gamma_1$ and to use it in order to map (push) one distribution on the other. To do this in practice, we consider the adjacency matrix $\mathcal{A} \in \mathbb{R}^{N\times N}$ associated with the graph $\mathcal{G}$ defined previously and notice that $\gamma_0$ (resp. $\gamma_1$) corresponds to those rows (nodes) in $\mathcal{A}$ for which $S=0$ (resp. $S=1$). Let us denote such submatrices of $\mathcal{A}$ by $\mathcal{A}_0 \in \mathbb{R}^{N_0\times N}$ and $\mathcal{A}_1 \in \mathbb{R}^{N_1\times N}$ and assume that they contain $N_0$ and $N_1$ rows, respectively. For further convenience, we denote the ratio of each class  by $\pi_s,\ s\in\{0,1\}$ with $\pi_s = \frac{N_s}{N}$. We now aim to solve the following OT problem:
\begin{align}
    \min_{\gamma \in \Pi(\frac{\bm{1}}{N_0},\frac{\bm{1}}{N_1})} \Omega_{\text{Group}}^{(\gamma,M)},\quad  \Omega_{\text{Group}}^{(\gamma,M)} := \langle \gamma, M\rangle
    \label{ot:group}
\end{align}
where $\frac{\bm{1}}{N_s}$ is a uniform vector with $N_s$ elements, $s \in \{0,1\}$ and $M$ is the matrix of pairwise distances between the rows in $\mathcal{A}_0$ and $\mathcal{A}_1$, \ie, $M_{ij} = l(a_{0}^{(i)},a_{1}^{(j)})$ for some distance $l$ where $a_{0}^{(i)}, a_{1}^{(j)}$ denote the $i$\textsuperscript{th} and $j$\textsuperscript{th} rows of $\mathcal{A}_0, \mathcal{A}_1$, respectively. Note that while the Hamming distance $l=\bm{1}_{\cdot\neq \cdot}$ is advocated by the theoretical results given above, in practice we use the usual squared Euclidean distance as it tends to give better results and allows a simple closed-form repairing procedure as detailed below.

\subsection{Individual fairness with Laplacian regularization}
Intuitively, one expects from an individually fair mapping to respect the initial relationships between the studied objects when learning their fair representation. This intuition of individually fair mapping was formally captured in the seminal work of \cite{DworkHPRZ12} and we present its adaptation to graphs below. 
\begin{definition} A mapping $\phi:\mathbb{V} \rightarrow \mathbb{V}$ satisfies the $(D,d)$-Lipschitz property if for every $v,v' \in \mathbb{V}$, and two metrics $D, d:\mathbb{V}\times \mathbb{V} \rightarrow \mathbb{R}_+$, we have
\begin{align*}
    D(\phi(v), \phi(v')) \leq d(v,v').
\end{align*}
\end{definition}
In our case, it means that for any two rows $a^{(i)}, a^{(j)} \in \mathcal{A}$, the adjacency matrix repaired by $\phi$ should preserve the initial similarities between them. We can further rewrite this definition in the form of constraints as follows:
$$D(\phi(a^{(i)}), \phi(a^{(j)})) k(a^{(i)},a^{(j)}) \leq 1,$$
where $k(a^{(i)},a^{(j)})$ is a similarity function inversely proportional to $d(a^{(i)},a^{(j)})$ (eg, as a popular RBF kernel). We can further assume that the mapping $\phi(\cdot)$ is given by a linear transformation with an unknown linear operator matrix $\Phi$. If $D$ is taken to be the squared $\ell_2$ norm, we get the following pair-wise constraints for all $a^{(i)}, a^{(j)}$:
$$||\Phi^Ta^{(i)}-\Phi^Ta^{(j)}||_2^2 k(a^{(i)},a^{(j)}) \leq 1.$$
One can further incorporate this term into the objective function by introducing a regularization term that aggregates pairwise constraints over all pairs:
$$\min_\Phi \sum_{i,j}||\Phi^Ta^{(i)}-\Phi^Ta^{(j)}||_2^2 k(a^{(i)},a^{(j)}).$$
We call this term ``individual fairness'' term, denote it by $\Omega_{\text{Indiv.}}$ and rewrite it as a graph Laplacian with similarity matrix $(K)_{ij} = k(a^{(i)},a^{(j)})$:
$$\Omega_{\text{Indiv.}}(\Phi,\mathcal{A},k) = \tr(\mathcal{A}^T\Phi^TL_K\Phi \mathcal{A}),$$
where $L_K = \diag(K) - K$.

Below, we put all the ingredients together to propose a unified repair procedure that allows us to control the extent of both group and individual fairness when solving the edge prediction task.

\subsection{Repairing the adjacency matrix}
The proposed optimization problem for both group and individually fair adjacency matrix repair takes the following form:
\begin{align}
\min_{\gamma \in \Pi(\frac{\bm{1}}{N_0},\frac{\bm{1}}{N_1})}\! \Omega_{\text{Group}}^{(\gamma,M)}\! + \! \lambda\sum_{i=0}^1 \Omega_{\text{Indiv.}}(\Phi_i(\gamma),\mathcal{A}_i,\textnormal{KNN}_3)
\label{OT:binary_repair}
\end{align}
where $\Phi_0(\gamma) = N_1\gamma^T$ and $\Phi_1(\gamma) = N_0\gamma$ are barycentric projections used to push the points of one distribution to those of the other \cite{ferradans}, and $\textnormal{KNN}_3$ is the adjacency matrix of a k-nearest neighbor graph with $k=3$ calculated from the raw adjacency matrix. Note that we choose to calculate the Laplacian using a KNN graph instead of the raw adjacency matrix as it provides a richer structural information about the graph. 
Once a solution $\gamma^{*}_\lambda$ to Problem \eqref{OT:binary_repair} is found, we use it to align the two joint conditional distributions by mapping both $\mathcal{A}_0$ and $\mathcal{A}_1$ on the mid-point of the geodesic path between them \cite{opac-b1129524} as follows:
\begin{align*}
    \tilde{\mathcal{A}_0} = \pi_0\mathcal{A}_0+\pi_1\gamma^{*}_\lambda\mathcal{A}_1,\quad \tilde{\mathcal{A}_1} &= \pi_1\mathcal{A}_1+\pi_0\gamma^{*T}_\lambda\mathcal{A}_0.
\end{align*}
Note that the closed-form expression given above is valid only when one uses the squared Euclidean distance between the nodes' representations. However, for any arbitrary distance we may obtain the equivalent solution by solving the pre-image problem for each row $\tilde{a}_{0}^{(i)}$ of $\tilde{\mathcal{A}_0}$, $i=\{1,\dots,N_0\}$ as follows
\begin{align*}
    \tilde{a}_{0}^{(i)} = \pi_0a_{0}^{(i)}+\pi_1\argmin{a \in \mathbb{R}^N}\sum_{j=1}^{N_0}\gamma^*_\lambda(i,j) l\left(a,a_{1}^{(j)}\right)
\end{align*}
and similarly for $\tilde{\mathcal{A}_1}$. Such an optimization procedure can be easily parallelized for all $\tilde{a}_{0}^{(i)} $ with each individual problem solved efficiently by any quasi-Newton method. 


\paragraph{Multi-class extension} In order to extend our method to the case of $|S|>2$, \ie, non-binary attributes, we propose to use a recently proposed method for computing the free-support Wasserstein barycenters introduced in \cite[Algorithm 2]{Cuturi14} and add a Laplacian regularization to it. This leads to the following optimization problem:
\begin{align*}
    \tilde{\mathcal{A}}_{\text{bary}} = \argmin{\mathcal{A}\in \mathbb{R}^{N\times N}}\frac{1}{|S|}\sum_{i=1}^{|S|}\min_{\gamma_i \in \Pi(\frac{\bm{1}}{N},\frac{\bm{1}}{N_i})} \Omega_{\text{Group}}^{(\gamma_i,M_i)}
    + \lambda\Omega_{\text{Indiv.}}(N_i\gamma_i^T, \mathcal{A},\textnormal{KNN}_3),
\end{align*}
where $M_i$ is the cost matrix between $\mathcal{A}$ and $\mathcal{A}_i$ for $i\in \{1,\dots, |S|\}$. Contrary to the binary setting of Problem \eqref{OT:binary_repair}, we have only one fairness term here applied to a projection of each sensitive group on the barycenter. Once the optimal solution for this problem is obtained, we use barycentric mapping to repair each individual submatrix $\mathcal{A}_i$ as follows:
\begin{align*}
    \tilde{\mathcal{A}}_{i} = N_i\gamma_i^{*T}\tilde{\mathcal{A}}_{\text{bary}},
\end{align*}
where $N_i = |S=i|$ and in general we do not require $N_i = N_j,\ i\neq j$. Note that contrary to the binary case, this mapping projects each matrix $\mathcal{A}_i$ on the barycenter and not on the mid-point of the geodesic path as before. 
\begin{figure}[!t]
    \centering
    \includegraphics[width=0.8\textwidth]{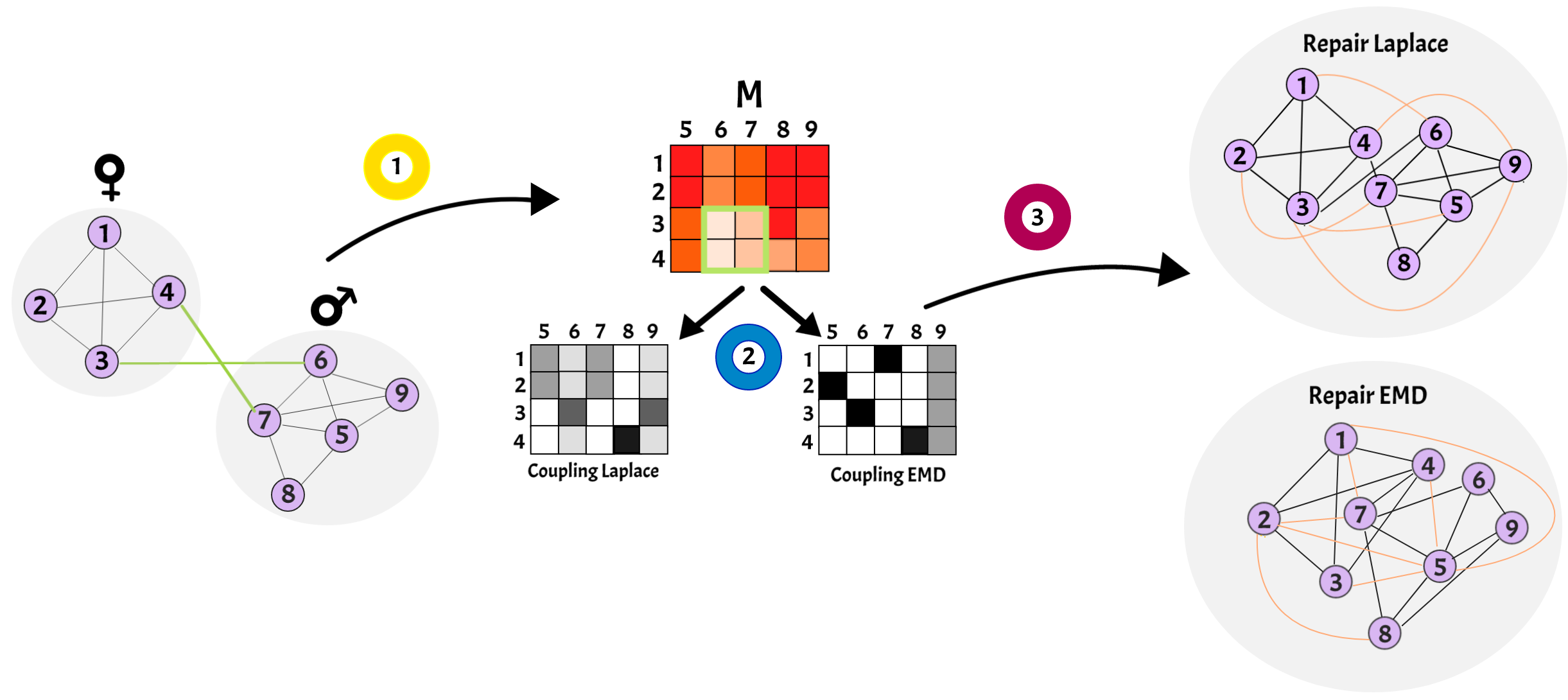}
    \caption{Illustration of the three steps performed to repair the adjacency matrix of a graph. 
    }
    \label{fig:illustration}
\end{figure}
\paragraph{Illustration} To illustrate the different steps needed to repair an adjacency matrix, we provide in Figure \ref{fig:illustration} a visual explanation of our proposed approach for a graph having 9 nodes ($\mathcal{A}$). In this graph, the nodes numbered from 1 to 4 belong to the class "Female" ($S=0$, $\mathcal{A}_0$), while the nodes from 5 to 9 belong to the class "Male" ($S=1$, $\mathcal{A}_1$). The matrix $M$ calculated in the \textbf{first step} and given in Figure \ref{fig:illustration} contains higher values (darker squares) for the node pairs that are far away from each other in terms of the used distance (e.g., 1 and 5) and lower values for those close to each other (e.g., 3 and 6). The solution obtained in the  \textbf{second step} highlights the difference between the group fair (EMD, $\lambda=0$) and both group and individually fair (Laplace, $\lambda>0$) repair as illustrated by the adjacency matrix $[\tilde{\mathcal{A}_0}\quad \tilde{\mathcal{A}_1}]$ obtained in the \textbf{third step}. Here, we see that group fair repair adds edges that obfuscate the original graph structure both within and across the sensitive groups, while adding individually fair regularization keeps the original structure withing groups almost intact. 

\subsection{Related works}
\paragraph{Fairness for graphs} To the best of our knowledge, very few articles proposed group fair repair schemes for relational data. In \cite{rahman2019}, the authors proposed \textsc{\textsc{Fairwalk}} algorithm that produces fairness-aware node embeddings using a modification of the random walk stage of the popular \textsc{\textsc{Node2vec}} algorithm \cite{GroverL16}. 
Similarly, \textsc{DeBayes} \cite{buyl2020} is an adaptation of Conditional Network Embedding (\textsc{Cne}) \cite{Kang2019CNE}, a Bayesian approach based on integrating prior knowledge through prior distribution for the network. 
\textsc{DeBayes} is an adaptation of \textsc{Cne} where the sensitive information is modeled in the prior distribution. Contrary to these two approaches, our proposal is embedding-agnostic, enjoys a theoretical justification and takes individual fairness into account as well.
Unlike the previous methods, \cite{bose19a} introduces an adversarial framework that enforces fairness by filtering out the information related to the sensitive attribute from node embeddings obtained with any embedding technique.  To obtain a trade-off between fairness and accuracy, the optimization process minimizes alternatively the loss w.r.t. the filtering and its opposite w.r.t. the discriminator. In that case, fairness is defined in terms of invariance, in other words independence, according to the mutual information, between the node embedding and the sensitive attribute.  A main drawback of such procedure is that it does not debias relational data given by pairs of nodes but only node embeddings themselves. Consequently, this algorithm seems more designed for fair node classification but not tailored to specifically tackle the fair edge prediction task that takes node tuples as input.

\paragraph{Fairness with OT} Several works used the capacity of OT to align probability distributions for fair classification \cite{Gordaliza19,Jiang2019,ZehlikeHW20}. The origin of such idea is close to the use of OT in domain adaptation \cite{CourtyFTR17} where two distributions are aligned using the barycentric mapping. Our work is close to this line of research and extends it in two ways. First, we show that Laplacian regularization previously used in OT for color transfer \cite{ferradans} and domain adaptation \cite{CourtyFTR17} leads to an individually fair repair. Second, we use the free-support barycenter algorithm to provide a multi-class version of repair that can deal with sensitive attributes taking non-binary values.

\section{Experimental evaluation}\label{sec:results}
We investigate the efficiency of our contribution at different levels for both synthetic graphs (see Appendix) and three real-world networks \footnote{The code reproducing the experimental results will be made publicly available upon acceptance of this paper.}. 
Overall, we aim to answer the following questions: 
\textbf{(Q1)}\textbf{ Impact on the structure of the graph}: we investigate the structural changes of the considered graph resulting from the repairing mechanism and the impact of the Laplacian regularization parameter used to promote individual fairness.
\textbf{(Q2)} \textbf{Impact on node embeddings}: we analyze the impact of our approach on the embeddings obtained from the repaired graph with traditional fairness-unaware node embedding methods. Specifically, we aim to verify whether one can infer the sensitive attribute from the node embedding vectors.
\textbf{(Q3}) \textbf{Impact on edge prediction}: we study the influence of the alterations of the graph on the edge prediction accuracy.

\subsection{Baselines}
For \textbf{(Q2)} and \textbf{(Q3)}, we consider two embedding methods, namely \textsc{Node2vec} (\textsc{N2Vec}) and \textsc{CNE}, and compare our approach with their fair versions described below.
\paragraph{\textsc{Node2vec} and \textsc{Fairwalk}}
To evaluate the impact of our approach on node embeddings, we first consider the very popular \textsc{\textsc{N2vec}} as a baseline. This approach builds a representation of a node in based on its neighborhood following a two-step procedure: 
\begin{enumerate}[wide, labelwidth=!, labelindent=0pt]
    \item Generate a corpus of traces by performing random walks. Formally, denoting by $c_i$ the $i$-th node in a given walk, the next node is selected among all neighbors of $c_i$, \ie, 
    $$\mathbb{P}(c_{i+1} = v|c_{i} = u) = \left\{
    \begin{array}{ll}
        \frac{\pi_{vu}}{C} & \mbox{if } \{u,v\} \in \mathcal{E} \\
        0 & \mbox{otherwise,}
    \end{array}
    \right.$$
    where $\pi_{uv}$ denotes the unnormalized transition probability between nodes $u$ and $v$ and $C$ corresponds to a normalization constant. The transition probability is set so as to reflect the neighborhood of $u$.
    \item Use the generated corpus to learn the embedding vectors through a SkipGram architecture that maximizes the log-probability of observing a network neighborhood for a node conditioned on its feature representation: 
    $$\argmax{Z}\prod_{u\in\mathcal{V}}\prod_{v \in N_u} \mathbb{P}(v|Z(u)).$$
\end{enumerate} 
We compare our approach with \textsc{\textsc{Fairwalk}}, a version of \textsc{\textsc{N2vec}}, designed for fair node embeddings. This latter modifies the transition probability of \textsc{\textsc{N2vec}} for the generation of unbiased traces. Step (1) of \textsc{\textsc{N2vec}} becomes 
$$\mathbb{P}(c_{i+1} = v|c_{i} = u) = \left\{
    \begin{array}{ll}
        \frac{1/k}{|S_{N_{u}}^{k}|} & \mbox{if } S_{v}^{k} = 1 \mbox{ and } \{u,v\} \in \mathcal{E} \\
        0 & \mbox{otherwise,}
    \end{array}
    \right.
$$
where $k = \{1,\cdots, K\}$ denotes the modality of the sensitive attribute $S$, $S_{N_{u}}^k$ is the number of nodes in the neighborhood of $u$ belonging to the group $k$ and $S_{v}^{k} = 1$ indicates that node $v$ belongs to the $k$-th group of the sensitive attribute. As a result, each generated random walk has a higher probability to contain nodes of different groups. 

\paragraph{\textsc{Cne} and \textsc{DeBayes}}
We also study the impact of our algorithm on \textsc{Cne} and compare our algorithm with its recently proposed fair extension \textsc{\textsc{DeBayes}}. 
Given a graph $G = (V, E, S)$, \textsc{\textsc{Cne}} finds an embedding $\textbf{Z}$ by maximizing $P(G|\textbf{Z}) =  \frac{P(\textbf{Z}|G)P(G)}{P(\textbf{Z})}$. In our experiments, we use the prior knowledge about the node degree modeled by the prior distribution $P(G)$ 
expressed by the following constraint:
\begin{equation}
\sum_{v \in V}P((v,v') \in E) = \sum_{v \in V} \mathds{1} \Big((v,v') \in \mathcal{E}\Big) .
\label{CNEeq1}
\end{equation}
\textsc{DeBayes} extends \textsc{Cne}, with a prior to model the sensitive attribute by replacing the constraint \eqref{CNEeq1} with:
\begin{equation*}
\sum_{v \in V_s}P((v,v') \in E~|~S(v) = s)= \sum_{v \in V_s} \mathds{1} \Big((v,v') \in \mathcal{E}\Big),
\end{equation*}
where $V_s = \{v | S(v) = s\}$. With this prior, debiased embeddings containing less information about sensitive information are obtained during the training step. Then, the debiased link predictions are computed using these embeddings and $P(G)$ instead of the biased prior distribution $P(G|S)$.

\paragraph{Random}
To illustrate the fact that the OT repairing schema is efficient in choosing where to add edges to reduce the bias, while maintaining a reasonable accuracy for link prediction, we also compare it with an approach that adds random edges between nodes from different groups for the sensitive attribute. This method is referred to as \textsc{Random}.

For the sake of reproducibility, all hyperparameters used in the experiments are provided in the Appendix. 

\subsection{Datasets}
We present the experimental results obtained on three real-world publicly available networks described below. Their key characteristics are summarized in Table \ref{tab:stat}. 

\textit{Political Blogs \cite{Adamic2005}}\footnote{ \url{www-personal.umich.edu/~mejn/netdata/}} is a network representing the state of the political blogosphere in the US in 2005. Nodes represent blogs and vertices represent hyperlinks between two blogs. For each node, the sensitive variable indicates the political leaning of the blog. 

\textit{Snap Facebook \cite{Leskovec2012}}\footnote{ \url{snap.stanford.edu/data/ego-Facebook.html}} data set consists of ego networks collected through the Facebook app. We use the combined version which contains the aggregated networks of ten individual's Facebook friends list. The gender is the sensitive attribute of each node. 

\textit{DBLP} is a co-authorship network originally built from DBLP, a computer science bibliography database. We use the version proposed by \cite{buyl2020} where the sensitive attribute corresponds to the continent extracted from authors' affiliation.  
\begin{table}
    \centering
    \caption{Statistics for all networks:  number of nodes ($\vert\mathcal{V}\vert$), number of edges ($\vert\mathcal{E}\vert$), type of the protected attribute.}
    \label{tab:stat}
    \begin{tabular}{c|c|c|c|c}
    Network & $\mathcal{V}$ &$\mathcal{E}$ & Type of $S$ &$\vert S\vert$\\
    \hline
    \textsc{PolBlogs}&$1,490$&$19,090$&binary&2\\
    \textsc{Facebook}&$4,039$&$88,234$&binary&2\\
    \textsc{DBLP}&3,790&6,602&multiclass&5\\
    \end{tabular}
\end{table}

\subsection{Experimental results}
\begin{table}
    \centering
     \caption{Comparison of assortativity coefficient w.r.t the protected attribute between the original and the repaired graphs. For the Laplacian, we report the values obtained for $\lambda \in \{0.005, 1, 5\}$}
    \label{tab:graph_structure}
    \begin{tabular}{c|c|c|c|c|c}
     Dataset    &Original&EMD&$\text{Lap}_{.005}$&$\text{Lap}_{1}$&$\text{Lap}_{5}$\\
    \hline
    \textsc{PolBlogs}    &$.81$&$.14$&$.59$&$.68$&$.77$ \\
    \textsc{Facebook}    &$.09$&$.04$&$.04$&$.05$&$.06$\\
    \textsc{DBLP}    &$.83$&$-.003$&$-.002$&$.04$&$.03$\\
    \end{tabular}
\end{table}
\begin{table*}[!ht]
    \centering
\caption{AUC score for link prediction, Representation Bias (RB), Disparate Impact (DI) and Consistency (Cons.). For the Laplacian, results corresponds to the regularization parameter set to 1. \textsc{Random} results are N2Vec-based.}\label{tab:edge_pred}
\resizebox{\textwidth}{!}{
 \begin{tabular}{l|c|cccc||cccc||c}
   &Metric&\textsc{N2Vec}&\textsc{\textsc{Fairwalk}}&\textsc{N2Vec$^{\text{EMD}}$}&\textsc{N2Vec$^{\text{Lap}}$}&\textsc{\textsc{Cne}}&\textsc{\textsc{DeBayes}}&\textsc{\textsc{Cne}$^\text{EMD}$}&\textsc{\textsc{Cne}$^{\text{Lap}}$}&\textsc{Random}\\
    \hline
  \multirow{4}{*}{\textsc{PolBlogs}}&AUC&$\mathbf{.75\pm{.01}}$&$\mathbf{.75\pm{.01}}$&$.66\pm{.01}$&$.73\pm{.01}$&$\mathbf{.93\pm{.01}}$&$.88\pm{.01}$&$.86\pm{.01}$&$.91\pm{.02}$&$.53\pm{.01}$\\
  &RB&$.97\pm{.01}$&$.96\pm{.01}$&$\mathbf{.78\pm{.01}}$&$.94\pm{.01}$&$.97\pm{.01}$&$\mathbf{.64\pm{.04}}$&$.73\pm{.03}$&$.94\pm{.04}$&$.63\pm{.01}$\\
  &DI&$.10\pm{.02}$&$.20\pm{.01}$&$.54\pm{.07}$&$.25\pm{.02}$&$.03\pm{.02}$&$.53\pm{.05}$&$\mathbf{.83\pm{.05}}$&$.19\pm{.03}$&$.43\pm{.02}$\\
  &Cons.&$.75\pm{.02}$&$.73\pm{.01}$&$.77\pm{.10}$&$\mathbf{.91\pm{.01}}$&$.89\pm{.01}$&$.89\pm{.01}$&$.90\pm{.01}$&$\mathbf{.93\pm{.01}}$&$.90\pm{.04}$\\
\hline
\hline
 \multirow{4}{*}{\textsc{Facebook}}&AUC&$\mathbf{.98\pm{.01}}$&$.85\pm.00$&$.96\pm{.00}$&$.96\pm{.00}$&$\mathbf{.99\pm{.01}}$&$\mathbf{.99\pm{.03}}$&$\mathbf{.99\pm{.01}}$&$.98\pm{.01}$&$.49\pm{.04}$\\
  &RB&$.64\pm{.01}$&$\mathbf{.61\pm.01}$&$\mathbf{.61\pm{.00}}$&$.63\pm{.00}$&$.58\pm{.02}$&$.57\pm{.02}$&$\mathbf{.54\pm{.03}}$&$.58\pm{.02}$&$.56\pm{.02}$\\
  &DI&$.80\pm{.01}$&$\mathbf{.83\pm.00}$&$.80\pm{.01}$&$.80\pm{.00}$&$.93\pm{.03}$&$.91\pm{.03}$&$.98\pm{.01}$&$\mathbf{.99\pm{.05}}$&$.84\pm{.02}$\\
  &Cons.&$.96\pm{.00}$&$.94\pm.00$&$.96\pm{.01}$&$.96\pm{.00}$&$\mathbf{.97\pm{.01}}$&$.96\pm{.00}$&$\mathbf{.97\pm{.01}}$&$\mathbf{.97\pm{.00}}$&$.89\pm{.01}$\\
    \hline
    \hline
  \multirow{4}{*}{\textsc{DBLP}}&AUC&$\mathbf
  {.98\pm{.01}}$&$\mathbf{.98\pm{.01}}$&$.78\pm{.03}$&$.81\pm{.04}$&$\mathbf{.98\pm{.01}}$&$\mathbf{.98\pm{.01}}$&$.77\pm{.03}$&$.82\pm{.05}$&$.54\pm{.01}$\\
  &RB&$.77\pm{.00}$&$.77\pm{.01}$&$\mathbf{.58\pm{.04}}$&$\mathbf{.58\pm{.02}}$&$.55\pm{.02}$&$\mathbf{.51\pm{.02}}$&$.52\pm{.01}$&$\mathbf{.51\pm{.02}}$&$.59\pm{.01}$\\
  &DI&$.14\pm{.01}$&$.14\pm{.01}$&$\mathbf{1.26\pm{.04}}$&$1.02\pm{.05}$&$.03\pm{.01}$&$.04\pm{.01}$&$\mathbf{1.29\pm{.04}}$&$.98\pm{.05}$&$.43\pm{.03}$\\
  &Cons.&$.91\pm{.01}$&$.91\pm{.01}$&$.93\pm{.02}$&$\mathbf{.95\pm{.01}}$&$.91\pm{.01}$&$.90\pm{.02}$&$.94\pm{.01}$&$\mathbf{.97\pm{.01}}$&$.86\pm{.01}$\\
\end{tabular}
}
\end{table*} 
\paragraph{Q1: impact on the graph structure} In order to gain insights on the structural changes resulting from the repairing with our OT-based method, we propose to look at the coefficient of assortativity given the sensitive attribute of the original graph and its repaired versions. We recall that assortativity coefficient takes values in the range between -1 and 1, and that its high values indicate a preference for nodes within the group to be connected with each other w.r.t. a given attribute. Therefore, in our context, one can see the assortativity w.r.t. the protected attribute as a measure of \textit{how much biased the graph structure is}, where values close to 1 indicate a strong bias. From Table \ref{tab:graph_structure}, we observe that \textsc{PolBlogs} and \textsc{DBLP} are strongly biased with assortativity coefficients close to 1, while \textsc{Facebook} presents no particular bias w.r.t. its protected attribute as indicated by the assortativity coefficient close to 0. Consequently, we expect 1) to significantly reduce this coefficient for \textsc{PolBlogs} and \textsc{DBLP} after the repair, and only slightly for \textsc{Facebook}, 2) to preserve the original bias more and more with the increasing strength of the Laplacian regularization. Both these expectations are confirmed by the results provided in Table \ref{tab:graph_structure} where the desired behavior is clearly observed. 

\paragraph{Q2: impact on node embeddings}
We proceed by studying the impact of the fair graph repair on the information carried by the node embeddings. In particular, we follow a standard protocol and use 10-fold cross-validated logistic regression to predict the sensitive attribute $S$ from the learned embeddings in order to understand whether applying these latter on a repaired graph maintains the desirable level of fairness. We use the resulting AUC score as a measure of bias, also termed Representation Bias (RB) in \cite{buyl2020}, and recall that in this context the ideal RB corresponds to the optimal value of $\text{BER}$ and should be around 0.5. These results are presented in Table \ref{tab:edge_pred}. From it, we can see that all repairing procedures manage to decrease the RB score successfully and that this decrease is more pronounced for \textsc{DeBayes} method and, in general, when using \textsc{Cne} embedding. We believe that this embedding is inherently more sensitive to the considered score and we leave the question on why this is the case as an open research avenue.   

\paragraph{Q3: Impact on edge prediction}
For \textsc{N2Vec}- and \textsc{Cne}-based approaches, we follow the protocol of \cite{rahman2019} and \cite{buyl2020} \footnote{\url{github.com/aida-ugent/DeBayes}}, respectively. Our goal here is to identify which approach provides the best trade-off in terms of fairness and prediction accuracy. 

\begin{figure}[!htpb]
    \centering
    \includegraphics[width=0.40\textwidth]{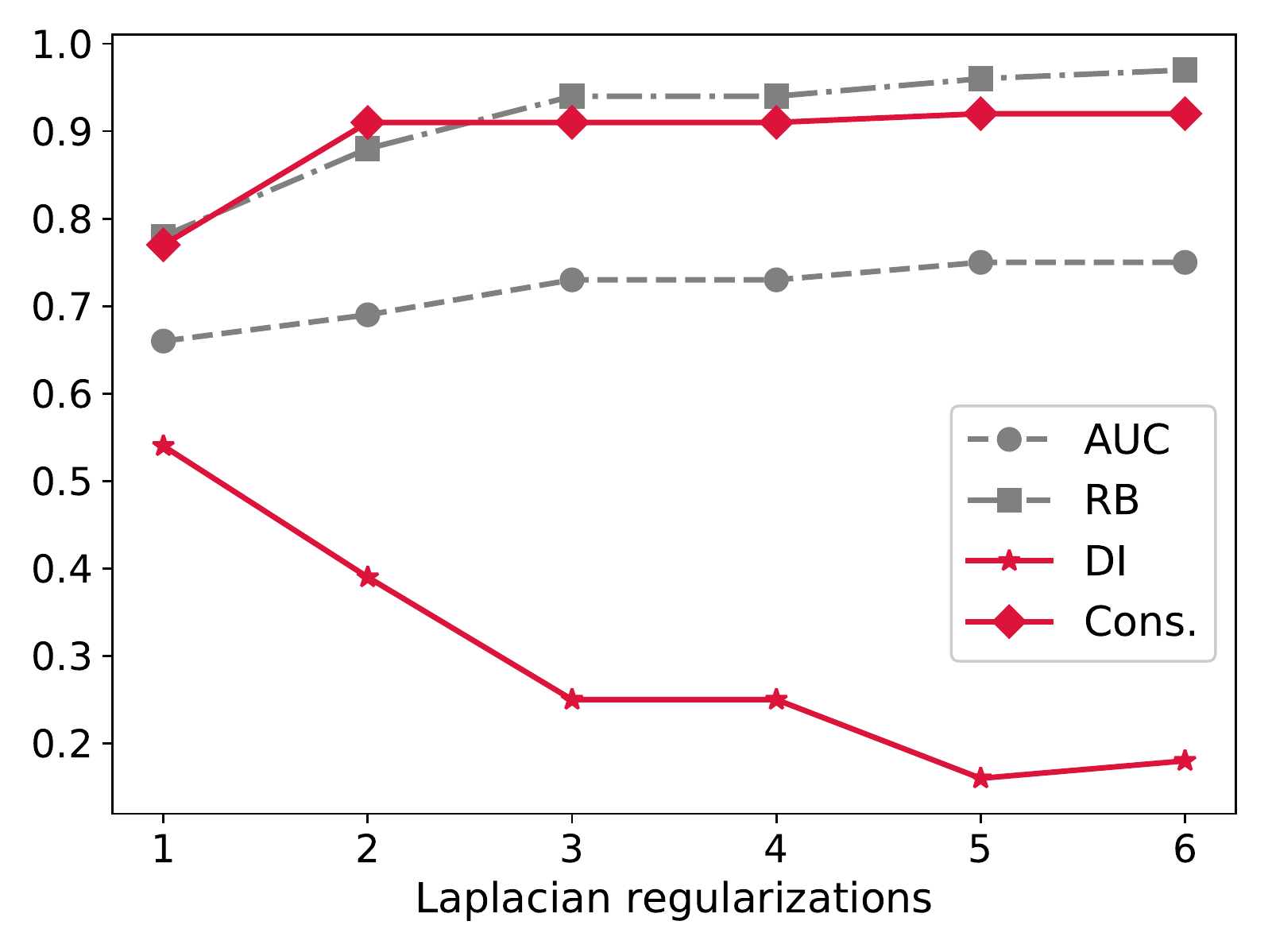}
    \caption{Impact of the Laplacian regularization on the different metrics for \textsc{Polblogs}.}
    \label{fig:reg}
\end{figure}

To this end, Table \ref{tab:edge_pred} reports the AUC for link prediction, the disparate impact (DI) and the consistency (Cons) scores, where the two latter are measures of group and individual fairness (see \cite{Zemel2013}), respectively. From these results, we make the following observations. First, we recall that \textsc{PolBlogs} and \textsc{DBLP} present a true challenge for fair edge prediction as the original results obtained with classical embeddings approaches are characterized by a low DI and a high RB score. This is contrary to \textsc{Facebook} graph embedding for which we obtain a high DI value indicating that it requires no particular repair. The goals of the repairing methods for each of these data sets are thus quite different: for \textsc{PolBlogs} and \textsc{DBLP} we would like to increase the DI value and reduce the predictability of the sensitive attribute by trading off as little of the edge prediction AUC as possible, while for \textsc{Facebook} the algorithms should mainly maintain the existing graph structure and not hinder the edge prediction with unnecessary repairing. 
From the obtained results, we first note that all fairness-aware methods increase DI score compared to the original one on \textsc{PolBlogs} and \textsc{Facebook}, while maintaining a decent prediction accuracy well-above the random guessing threshold observed in the case of the \textsc{Random} repair. On the other hand, on \textsc{DBLP} dataset only our approach improves fairness scores, but this comes at a price of a drop in terms of the performance. Most likely, this is due to the imbalance between different sensitives groups that hinders the performance of OT. 
As for the consistency, only Laplacian regularization significantly improves this criterion, while it remains almost unchanged for other baselines after the repair. While different repair methods have their distinct strong sides making it difficult to choose the ``best" one, we note that our proposed approach is versatile and allows to explicitly control the trade-off between the fairness and prediction accuracy and to be used with different embeddings. Finally, Figure \ref{fig:reg} shows the results for different regularization parameters for the Laplacian OT on \textsc{Polblogs}. Once can see that as the value of the regularization parameter increase, the group fairness metric (DI) decreases while the individual fairness metric (Cons.) increases. 



\section{Conclusion}
In this paper we addressed an important problem of fair edge prediction in graphs. Contrary to fair classification, fair edge prediction in graphs has received a very limited amount of attention from the research community and has mainly been solved using heuristic embedding-dependent procedures and only in group fairness context. To bridge this gap, we provide a first embedding-agnostic repair procedure for the adjacency matrix of a graph with both group and individual fairness constraints. We show through extensive experimental evaluations that our approach provides a flexibility of choosing explicitly to which extent one wants to ensure group and individually fair constraints. 

Further research directions of this work are many. First, we would like to study the impact of different embedding techniques on the bias in the adjacency matrix of a graph as empirical evidence suggests that some embedding techniques reinforce the bias in the data making it even more apparent. We also plan to use a recent theoretical analysis of popular node embedding methods \cite{QiuDMLWT18} to provably show their effect on the correlation between the estimated output and the sensitive attribute. 

\newpage

\Large{\hypertarget{appendix}{\textbf{Appendix}}}%

\normalsize
\setcounter{section}{0}
\def\thesection{\Alph{section}}

\section{Proofs}
\paragraph{Proof of Theorem 1}
\begin{proof}
We start by upper-bounding $\mathbb{P}(h(V,V')=1|S =1)$ and then proceed by giving a lower bound for $\mathbb{P}(h(V,V')=1|S =0)$. 
\begin{align}
    &\mathbb{P}(h(V,V')=1|S =1) = \frac{\mathbb{P}(S = 1|h(V,V')= 1)\mathbb{P}(h(V,V')= 1)}{\mathbb{P}(S = 1)}\label{l1}\\
    & = \frac{\mathbb{P}(S = 1, S' = 1 \vee S = 1, S' = 0|h(V,V')= 1)\mathbb{P}(h(V,V')= 1)}{\mathbb{P}(S = 1)}\label{l2}\\
    & \leq \frac{\mathbb{P}(S = 1, S' = 1 \vee S = 0, S' = 0|h(V,V')= 1)\mathbb{P}(h(V,V')= 1)}{\mathbb{P}(S = 1)}\label{l3}\\
    & = \frac{\mathbb{P}(S = S'|h(V,V')= 1)\mathbb{P}(h(V,V')= 1)}{\mathbb{P}(S = S')}\label{l4}\\
    &= \mathbb{P}(h(V,V')= 1|S = S').\notag
\end{align}
where \eqref{l1} is obtained from Bayes' theorem; \eqref{l2} follows from the fact that $P(A) = P(A|B)+P(A|\bar{B})$; \eqref{l3} and \eqref{l4} follow from assumptions A1 and A2, respectively.

Similarly, we get for $\mathbb{P}(h(V,V')=1|S =0)$:
\begin{align*}
    &\mathbb{P}(h(V,V')=1|S = 0) = \frac{\mathbb{P}(S = 0|h(V,V')= 1)\mathbb{P}(h(V,V')= 1)}{\mathbb{P}(S = 0)}\\
    & = \frac{\mathbb{P}(S = 0, S' = 1 \vee S = 0, S' = 0|h(V,V')= 1)\mathbb{P}(h(V,V')= 1)}{\mathbb{P}(S = 0)}\\
    & \geq \frac{\mathbb{P}(S = 0, S' = 1 \vee S = 1, S' = 0|h(V,V')= 1)\mathbb{P}(h(V,V')= 1)}{\mathbb{P}(S = 0)}\\
    & = \frac{\mathbb{P}(S \neq S'|h(V,V')= 1)\mathbb{P}(h(V,V')= 1)}{\mathbb{P}(S \neq S')}\\
    &= \mathbb{P}(h(V,V')= 1|S \neq S').
\end{align*}
Combined together, we obtain that:
$$\frac{\mathbb{P}(h(V,V')= 1|S \neq S')}{\mathbb{P}(h(V,V')= 1|S = S')} \leq \frac{\mathbb{P}(h(V,V')=1|S =0)}{\mathbb{P}(h(V,V')=1|S =1)} \leq \tau.$$
\end{proof}

\paragraph{Proof of Corollary 1}
\begin{proof}
We use Theorem 1 to obtain 
\begin{align*}
    &\text{DI}(h,\mathbb{V},S\oplus S') \leq \text{DI}(h,\mathbb{V},S) \leq \tau \\
    &\implies \frac{1}{\tau}\mathbb{P}(h(V,V')= 1| S \neq S') \leq 1 - \mathbb{P}(h(V,V')= 0| S = S')\\
    &\implies (1\!+\!(\frac{1}{\tau}\!-\!1))\mathbb{P}(h(V,V')\!=\! 1| S\! \neq \!S')\! + \! \mathbb{P}(h(V,V')\!= \! 0| S \!=\! S')\! \leq 1 \! \\
    &\implies \mathbb{P}(h(V,V')= 1| S \neq S') + \mathbb{P}(h(V,V')= 0| S = S')\\
    &\hspace{4cm} \leq (\frac{1}{\tau}-1)\mathbb{P}(h(V,V')= 1| S \neq S')\\
    &\implies \text{BER}(h,\mathbb{V},S\oplus S') \leq \frac{1}{2} - \frac{\mathbb{P}_1(h)}{2}\left(\frac{1}{\tau}-1\right).
\end{align*}
For the second part of the statement, we use \cite[Theorem 2.2]{Gordaliza19} to obtain 
$$\min_{h \in \mathcal{H}} \text{BER}(h,\mathbb{V},S\oplus S') = \frac{1}{2}(1-\text{d}_\text{TV}(\gamma_0,\gamma_1)).$$
We further use the equality between the Wasserstein distance with Hamming distance used as a cost function and the total variation metric to obtain the final result. 
\end{proof}

\section{Additional experiments and results}
\subsection{Experiments on synthetic networks}
We evaluate our approach on several synthetic graphs of controlled complexity. This allows us to cover different possible scenarios for the presence of bias in graphs in order to highlight the different features of our algorithm. 
\paragraph{Graph generation}
We generate five synthetic graphs (\Gu--\Gc) composed of 150 nodes each on the basis of the stochastic block model with different block parameters. This latter is done using the \texttt{networkx} library \footnote{\url{https://networkx.github.io/}}.
For the sake of reproducibility, we provide the parameters used to generate the graphs in Table \ref{tab:parameters}. We further assume that each node is associated with one value of sensitive attribute $S \in \{0, \cdots, K\}$. For the first four graphs we assume a binary sensitive attribute, while for {\Gc} we assume $S$ to be multiclass (K=3). 

\begin{figure}[!ht]
    \centering
    \subfloat[\Gu]{\includegraphics[width=0.30\textwidth]{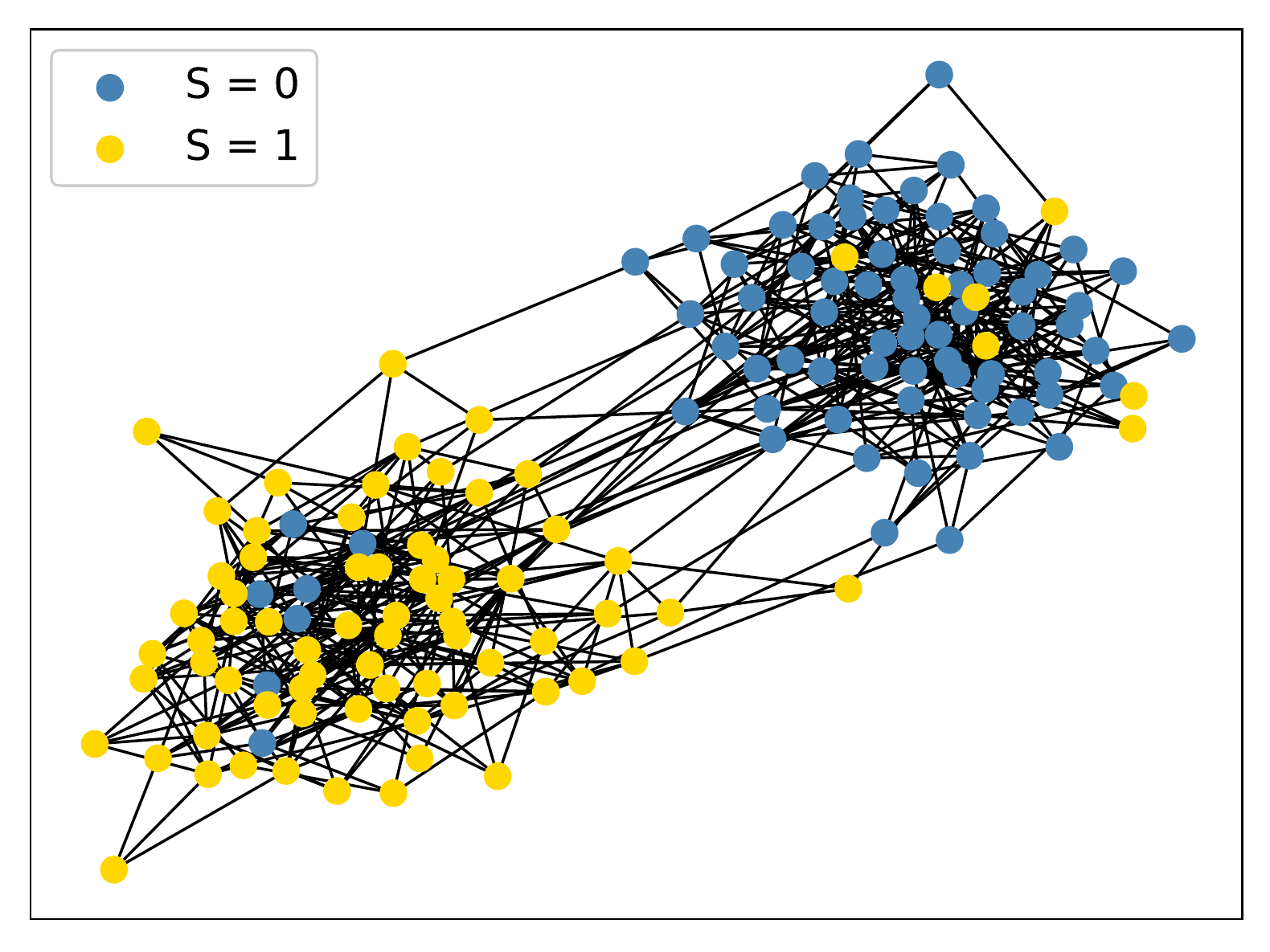}}
    \subfloat[\Gd]{\includegraphics[width=0.30\textwidth]{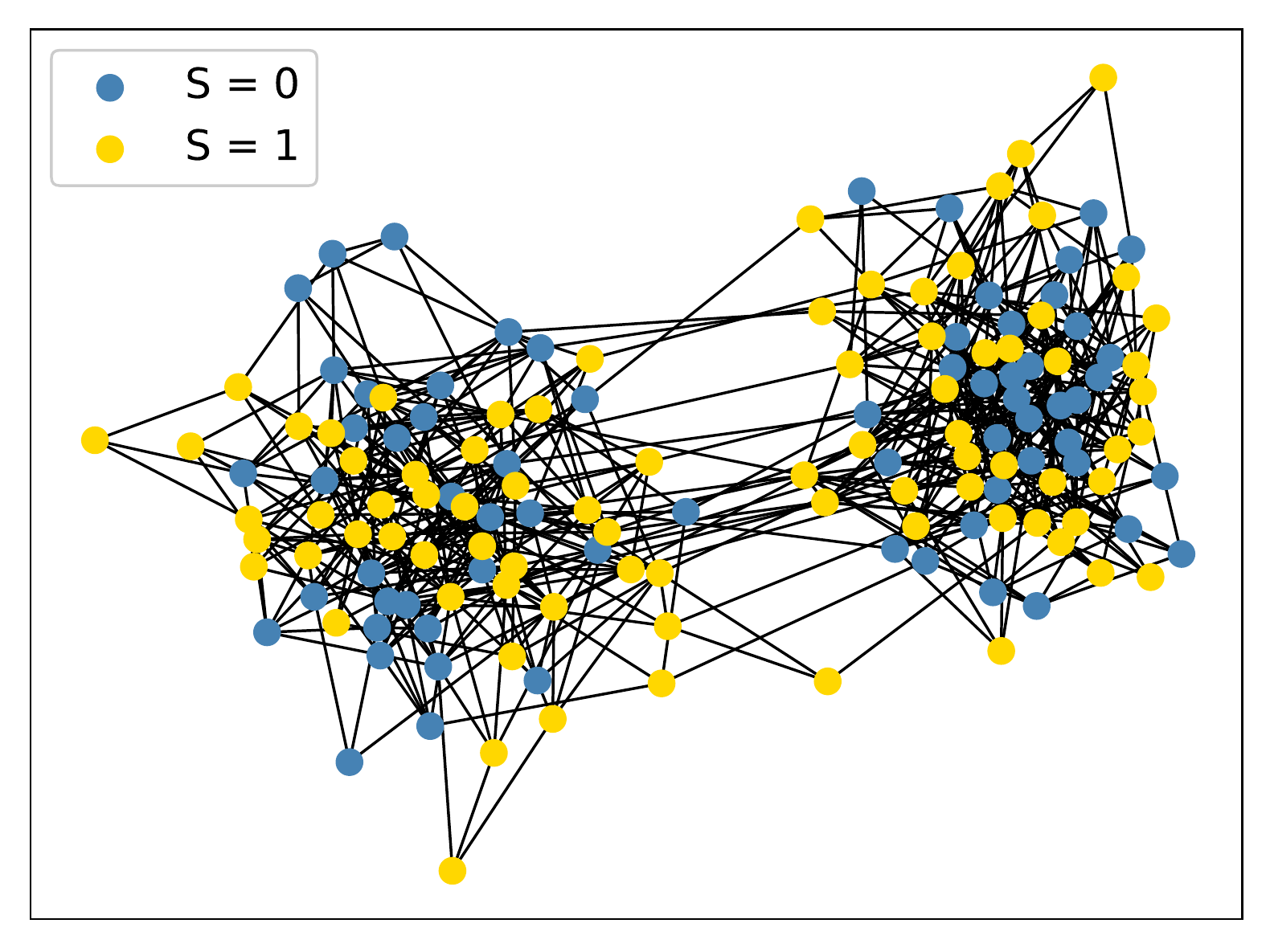}}
    \subfloat[\Gt]{\includegraphics[width=0.30\textwidth]{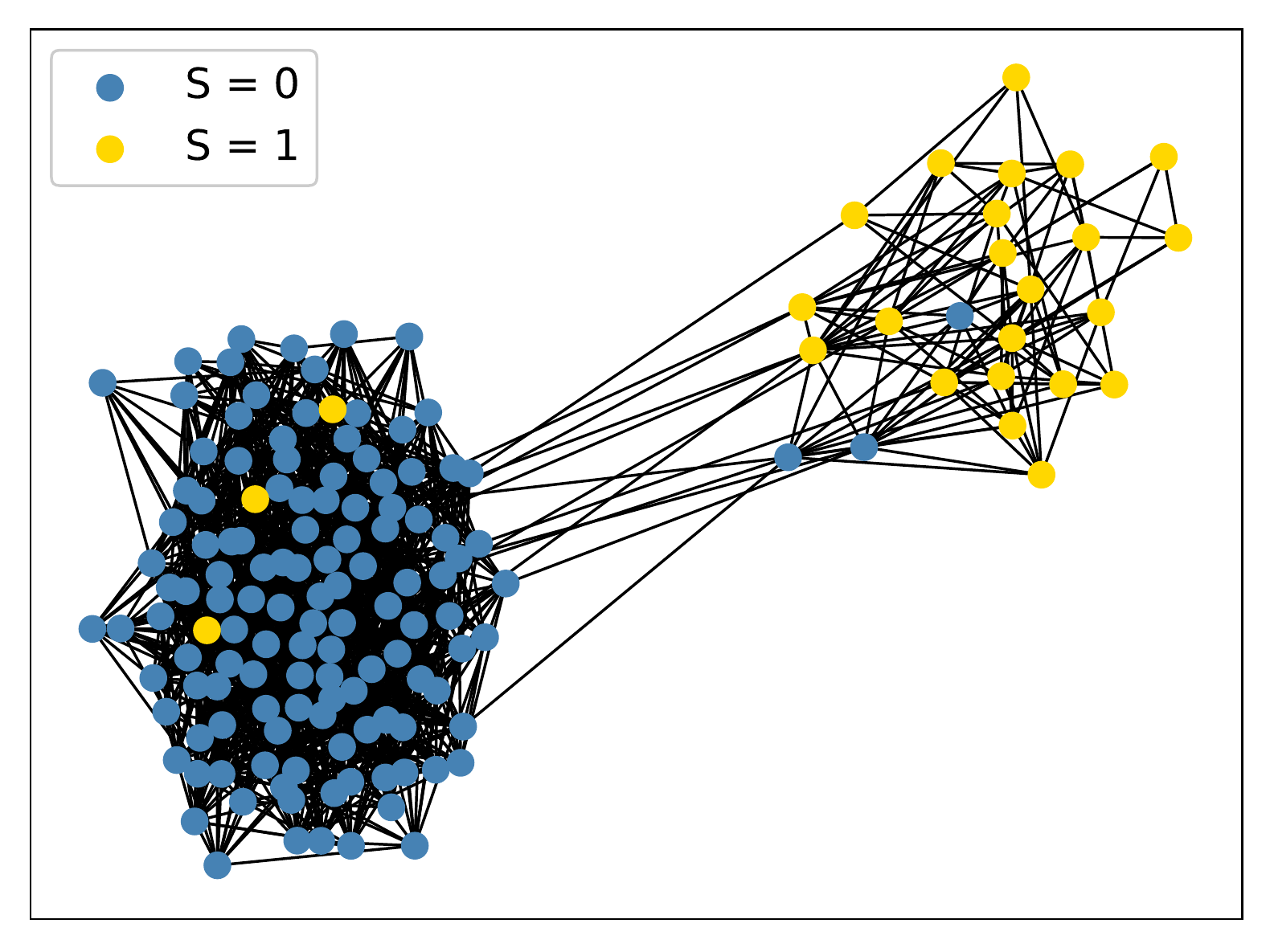}}\\
    \subfloat[\Gq]{\includegraphics[width=0.30\textwidth]{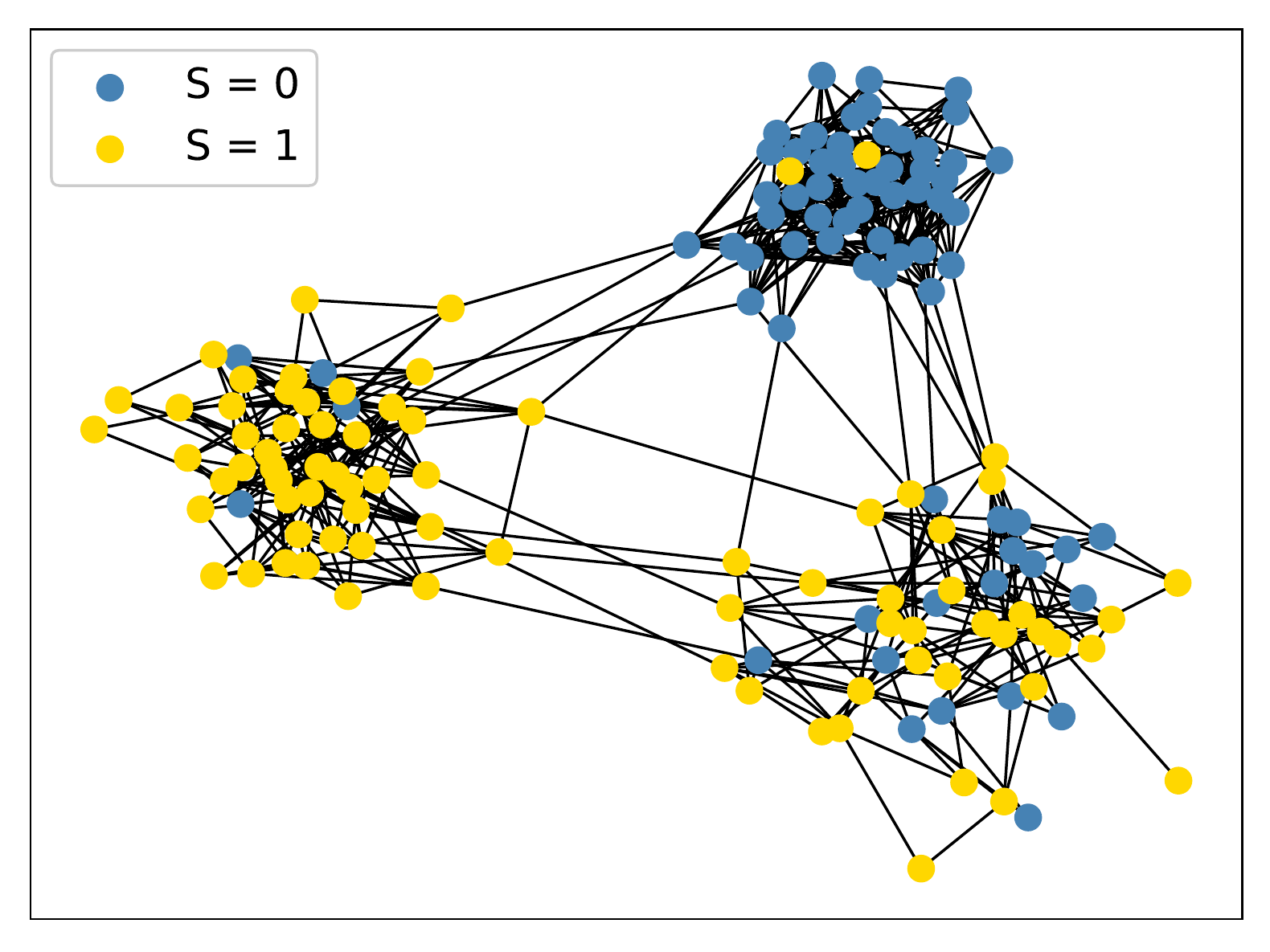}}
    \subfloat[\Gc]{\includegraphics[width=0.30\textwidth]{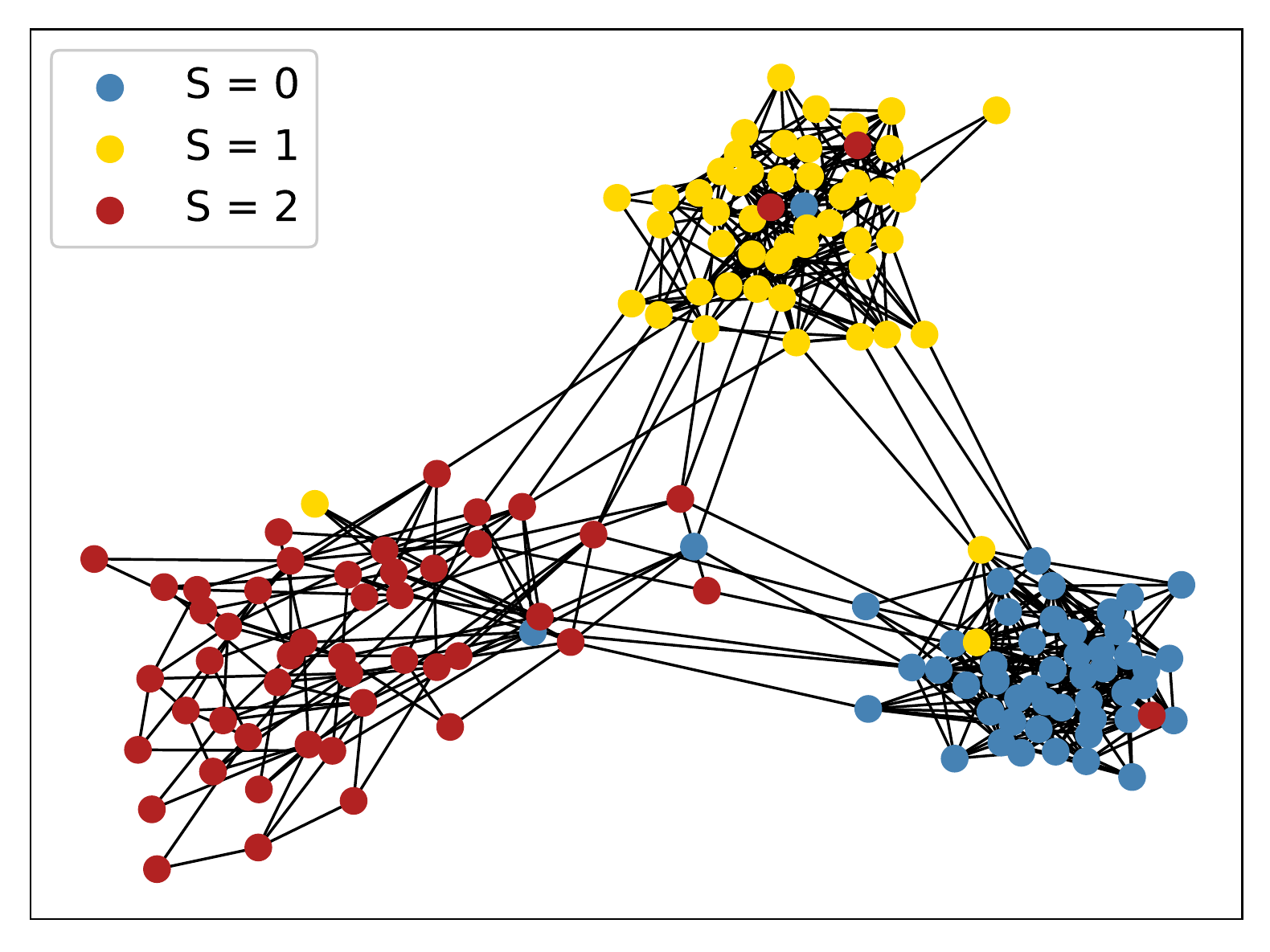}}
    \caption{Visualisation of the five synthetic graphs {\Gu}--{\Gc}. Node color indicates the value of the sensitive attribute associated with each node.}
    \label{fig:visu_synthetic}
\end{figure}

In terms of the structure, {\Gu} corresponds to a graph with two communities and a strong dependency between the community structure and the sensitive attribute $S$; {\Gd} has the same community structure as {\Gu}, but with $S$ being independent of the structure; {\Gt} corresponds to a graph with two imbalanced communities dependent on $S$ and a stronger intra-connection in the smaller community; {\Gq} is a graph with three communities, two of them being dependent on $S$, and the third one being independent. Finally, {\Gc} is similar to {\Gq} and has three communities, each of them being dependent on one of the class of the protected attribute. In the following, each graph is generated 50 times and the average metrics are reported. Figure \ref{fig:visu_synthetic} shows the different graphs.

\begin{table}[!htpb]
\caption{Parameters used for graph generation. \textit{Cluster} implies that the value of $S$ is almost equal to the community identifier, while \textit{random} implies that $S$ is randomly generated and is independent from the community structure of the graph.}
\label{tab:parameters}
    \centering
      \resizebox{0.6\textwidth}{!}{
    \begin{tabular}{ccccccc}
    \hline
    Graphs & $S$& $|S|$& Size & Probs\\ 
    \hline
     {\Gu}   & cluster &2& $\begin{pmatrix}75 &75 \end{pmatrix}$& $\begin{pmatrix}
      .10 &  .005 \\
      .005 & .10 \end{pmatrix}$\\
     {\Gd}   & random &2& $\begin{pmatrix}75 &75 \end{pmatrix}$& $\begin{pmatrix}
      .10 &  .005 \\
      .005 & .10 \end{pmatrix}$ \\
     {\Gt}   & cluster &2& $\begin{pmatrix}125 &25 \end{pmatrix}$&$\begin{pmatrix}.15& .005 \\ .005 &.35\end{pmatrix}$\\
     {\Gq}   & random \& cluster & 2&$\begin{pmatrix}50 &50 &50 \end{pmatrix}$& $\begin{pmatrix}.20& .002& .003\\ .002 & .15&.003\\.003 & .003 & .10\end{pmatrix}$ \\
     {\Gc}   & cluster &3& $\begin{pmatrix}50 &50 &50 \end{pmatrix}$& $\begin{pmatrix}.20& .002& .003\\ .002 & .15&.003\\.003 & .003 & .10\end{pmatrix}$ \\
    \hline
    \end{tabular}
    }
\end{table}

\begin{figure}[!htpb]
    \centering
    \subfloat[{\Gu}]{
    \includegraphics[width=0.47\textwidth]{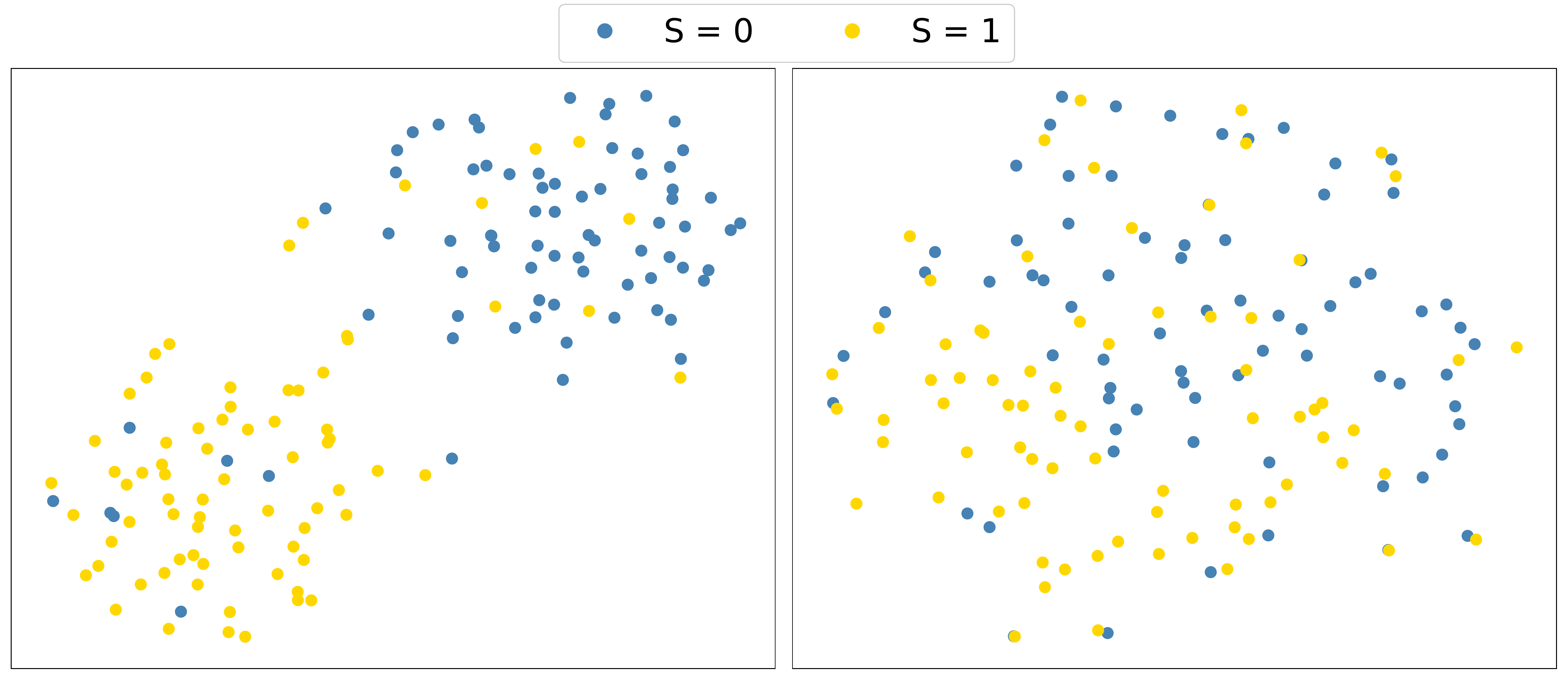}}
    \subfloat[{\Gd}]{
    \includegraphics[width=0.47\textwidth]{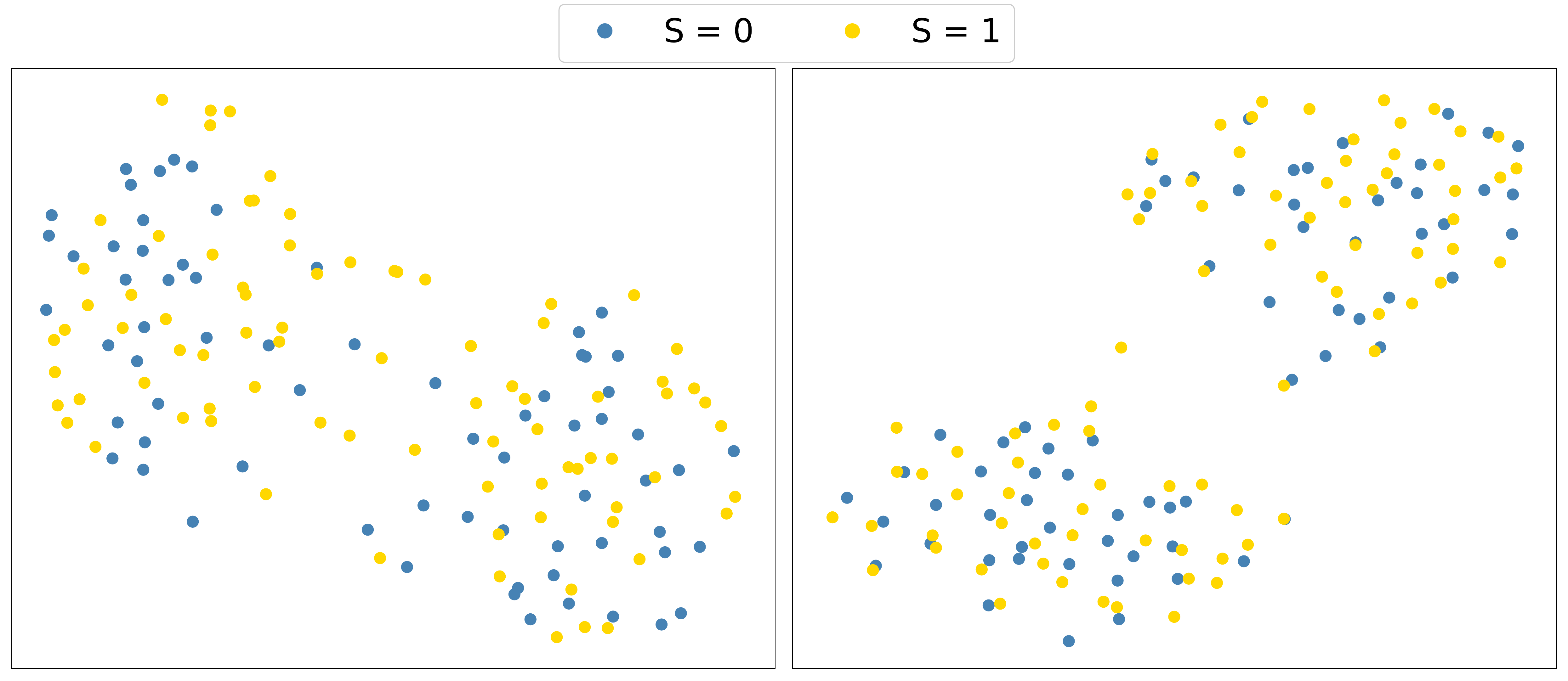}}\\
    \subfloat[{\Gt}]{
    \includegraphics[width=0.47\textwidth]{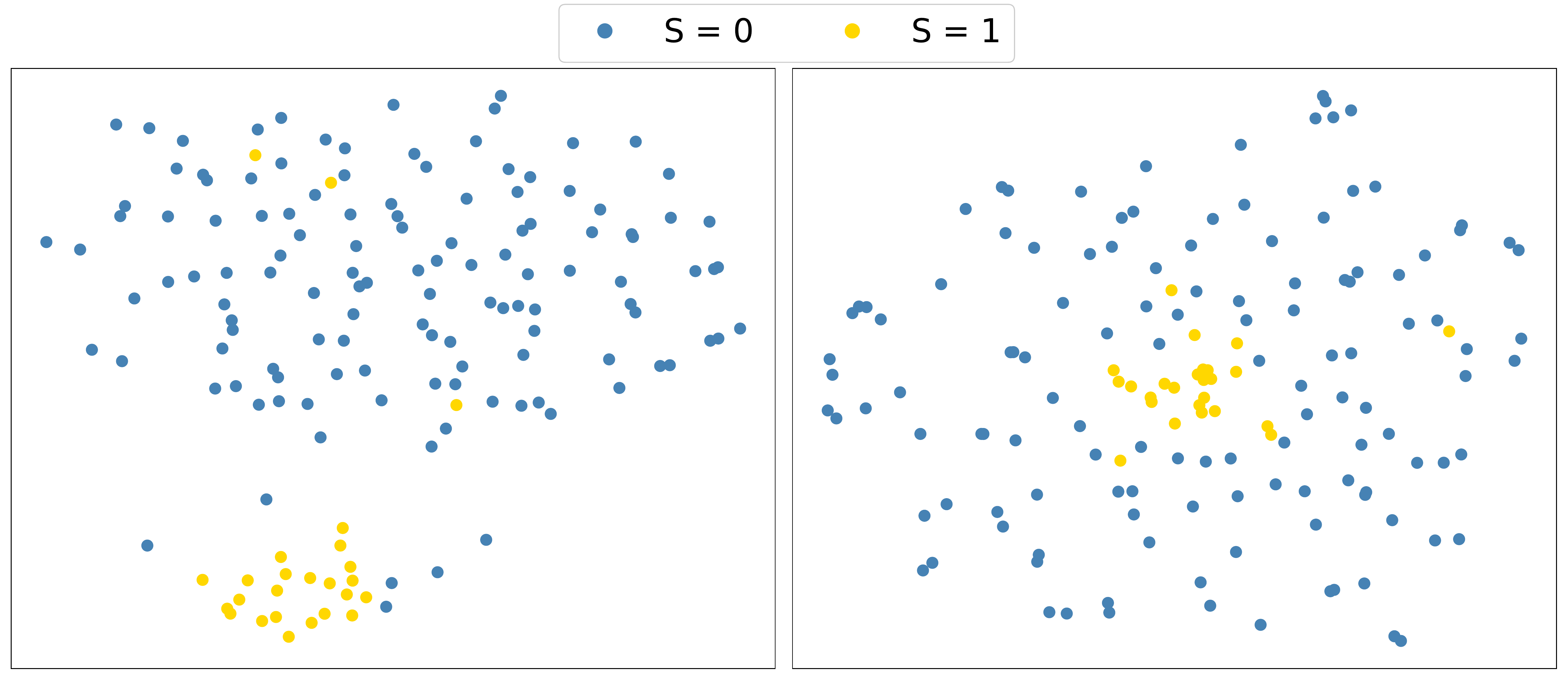}}
    \subfloat[{\Gq}]{
    \includegraphics[width=0.47\textwidth]{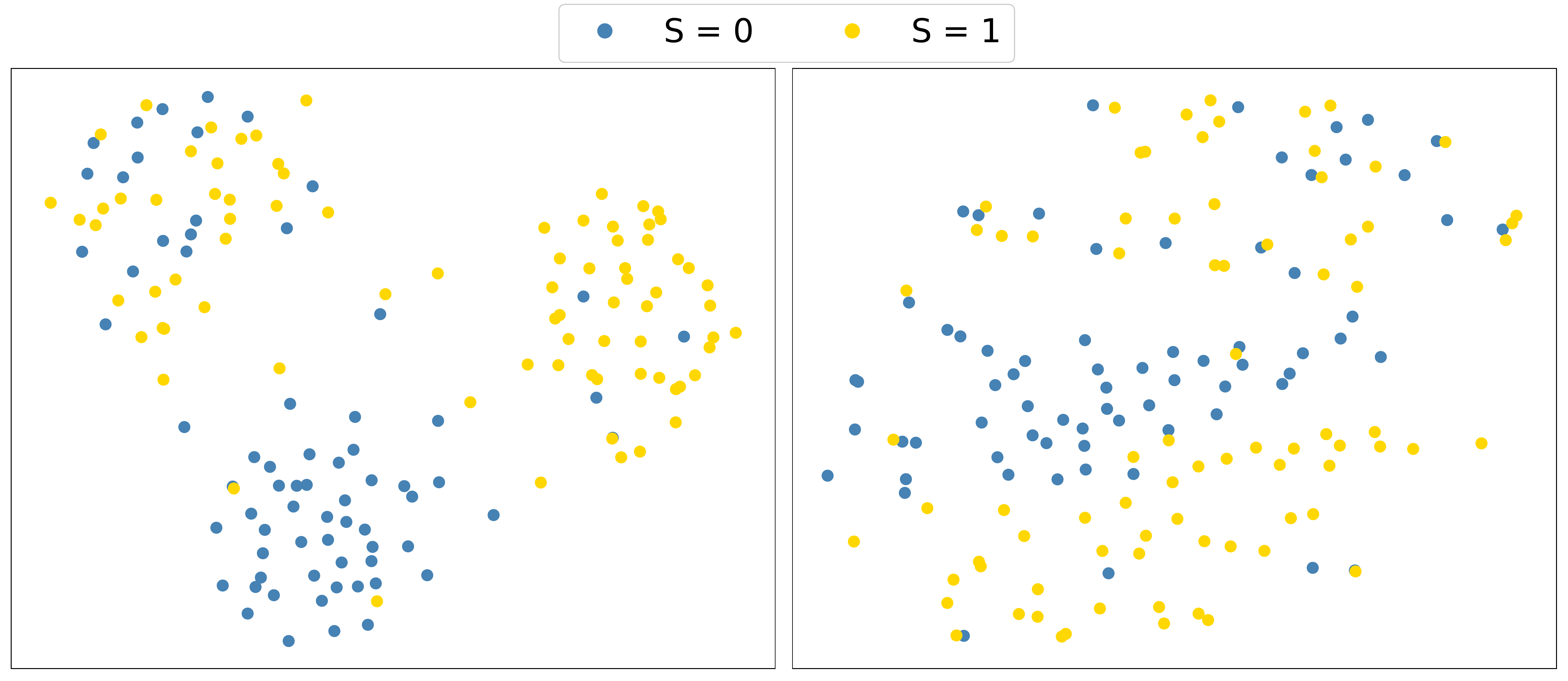}}\\
    \subfloat[{\Gc}]{
    \includegraphics[width=0.47\textwidth]{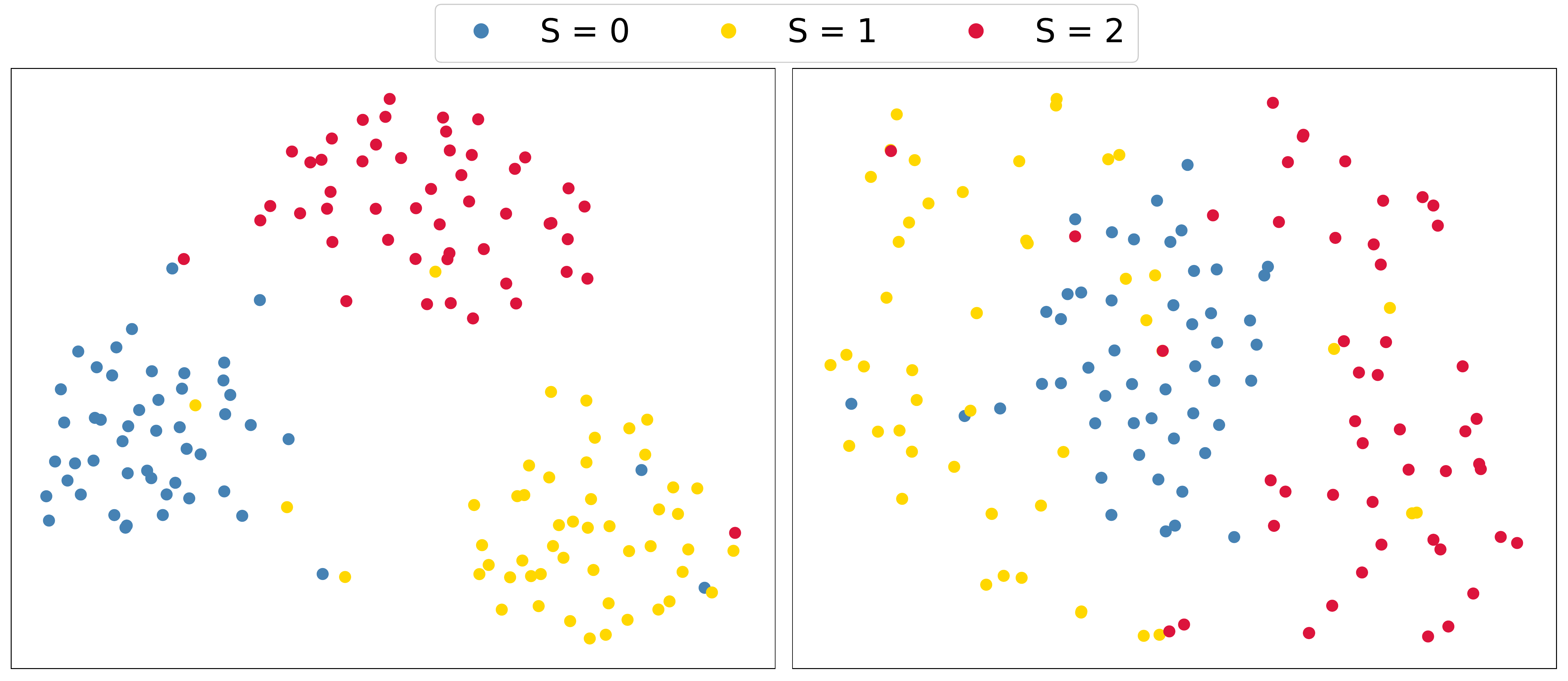}}
    \caption{Visualisation with t-SNE of node embeddings computed with \textsc{Node2Vec} on (left) the original graph and (right) on the repaired version.}
    \label{fig:tsne}
\end{figure}

\paragraph{Obtained results}
We now focus on the impact of the repairing and the Laplacian regularization on the graph topology and on the node embeddings generated with \textsc{N2Vec}. 
Table \ref{tab:graph_properties} reports the results for the EMD repairing on all graphs, which corresponds to the case without the individual fairness term. The assortativity coefficient is calculated conditionally on the sensitive attribute, and we recall that a value close to 1 (resp. -1) implies that the graph contains more edges between the nodes with the same (resp. different) value of protected attribute, while a value close to 0 implies that the distribution of edges is balanced between nodes belonging to the same and different groups. For the representation bias (RB), it corresponds to the AUC obtained by a classifier where the input is the node embedding and the output is the protected attribute. In our context, it should ideally be close to 0.5, \ie, the performance of random guessing. We also visualise the obtained embeddings in Figure \ref{fig:tsne}.

\begin{table}[!htpb]
\caption{Coefficients of assortativity conditionally on the protected attribute (Ass.) and representation bias (RB) for the original (O) and the repaired (R) graphs when using the EMD repairing (i.e. regularization is set to 0).}
\label{tab:graph_properties}
\centering
      \resizebox{\textwidth}{!}{
    \begin{tabular}{c|cc|cc|cc|cc|cc}
    \hline 
      &\multicolumn{2}{c|}{{\Gu}} &  \multicolumn{2}{c|}{{\Gd}}& \multicolumn{2}{c|}{{\Gt}}& \multicolumn{2}{c|}{{\Gq}}& \multicolumn{2}{c}{{\Gc}}   \\
    \cline{1-11}
     &\textsc{O} & \textsc{R} & \textsc{O} &\textsc{R}&\textsc{O} &\textsc{R}&\textsc{O} &\textsc{R}&\textsc{O} &\textsc{R}\\
    \hline
    Ass. &$.74\pm{.07}$&$.01\pm{.08}$&$-.01\pm{.04}$&$-.02\pm{.04}$&$.71\pm{.11}$&$-.03\pm{.02}$&$.60\pm{.07}$&$.06\pm{.11}$&$.74\pm{.07}$&$.32\pm{.06}$\\
    RB &$.92\pm{.04}$&$.48\pm{.06}$&$.52\pm{.06}$&$.44\pm{.06}$&$.96\pm{.04}$&$.43\pm{.08}$&$.85\pm{.04}$&$.55\pm{.11}$&$.95\pm{.02}$&$.91\pm{.03}$\\
    \hline
    \end{tabular}
    }
\end{table}

From Table \ref{tab:graph_properties}, one can see that the coefficient of assortativity and RB are initially high for graphs {\Gu}, {\Gt} and {\Gq}, implying that these graphs are strongly biased. In all cases, we observe that the assortativity values are much lower on the repaired graphs than on the original ones, indicating that $S$ is no longer correlated with their latent community structure. As for the RB, we also observe an important drop of the values after the repairing, indicating that the sensitive attribute can no longer be inferred from the embeddings learnt on the graph. For {\Gd}, however, both quantities of interest remain the same before and after the repair. This is explained by the fact that this graph is not biased and thus need no repair. Finally, in case of {\Gc}, the assortativity coefficient decreases after the repair, while RB remains high. We explain this by the fact that the used embedding technique manages to pick up the signal in the case of multi-class repair more easily and reintroduces the bias back to the task at hand. All these observations are confirmed by the plots of the repaired embeddings given in Figure \ref{fig:tsne}.

\begin{figure}[!htpb]
    \centering
    \subfloat[\Gu]{\includegraphics[width=0.30\textwidth]{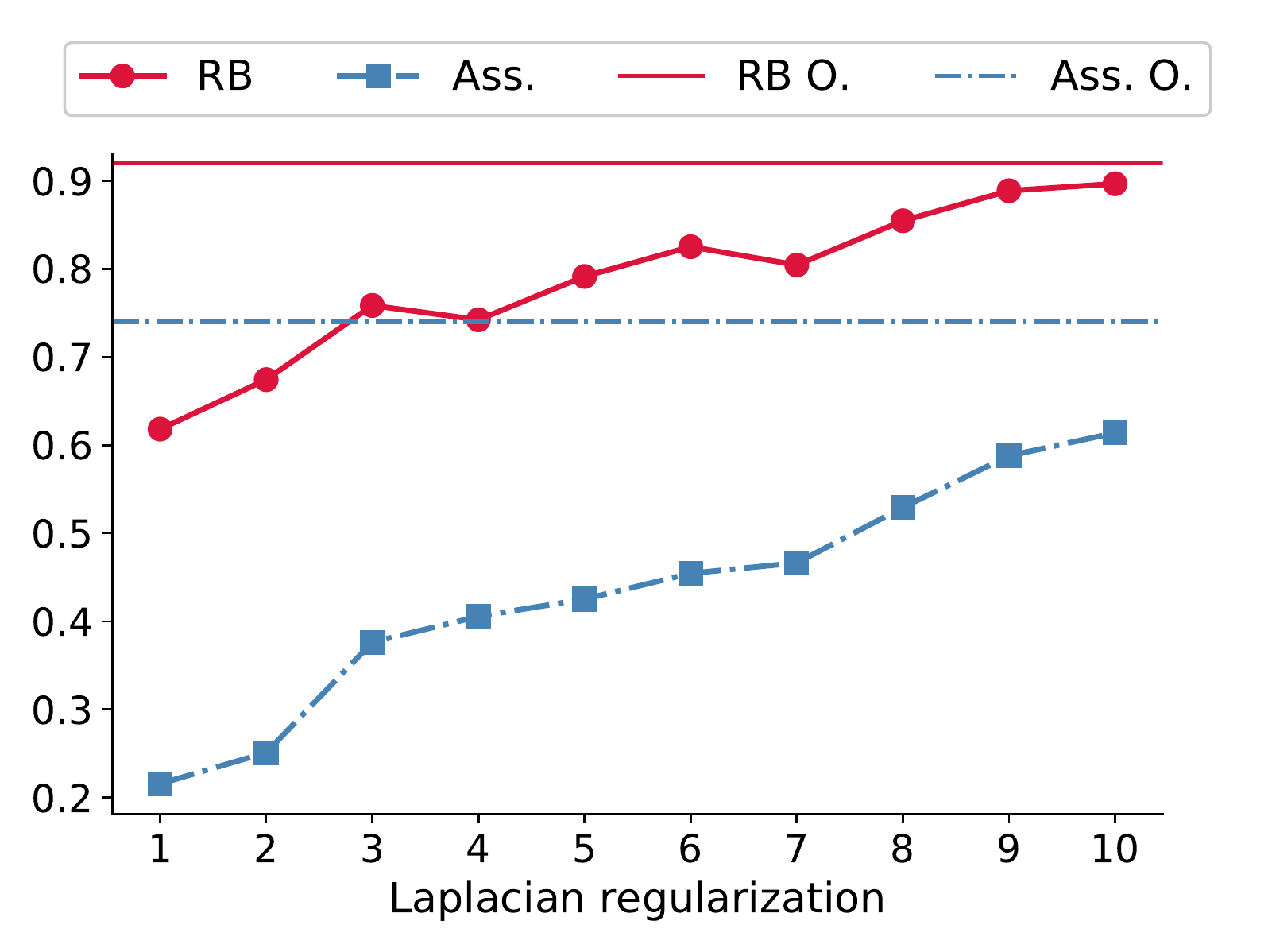}}
    \subfloat[\Gd]{\includegraphics[width=0.30\textwidth]{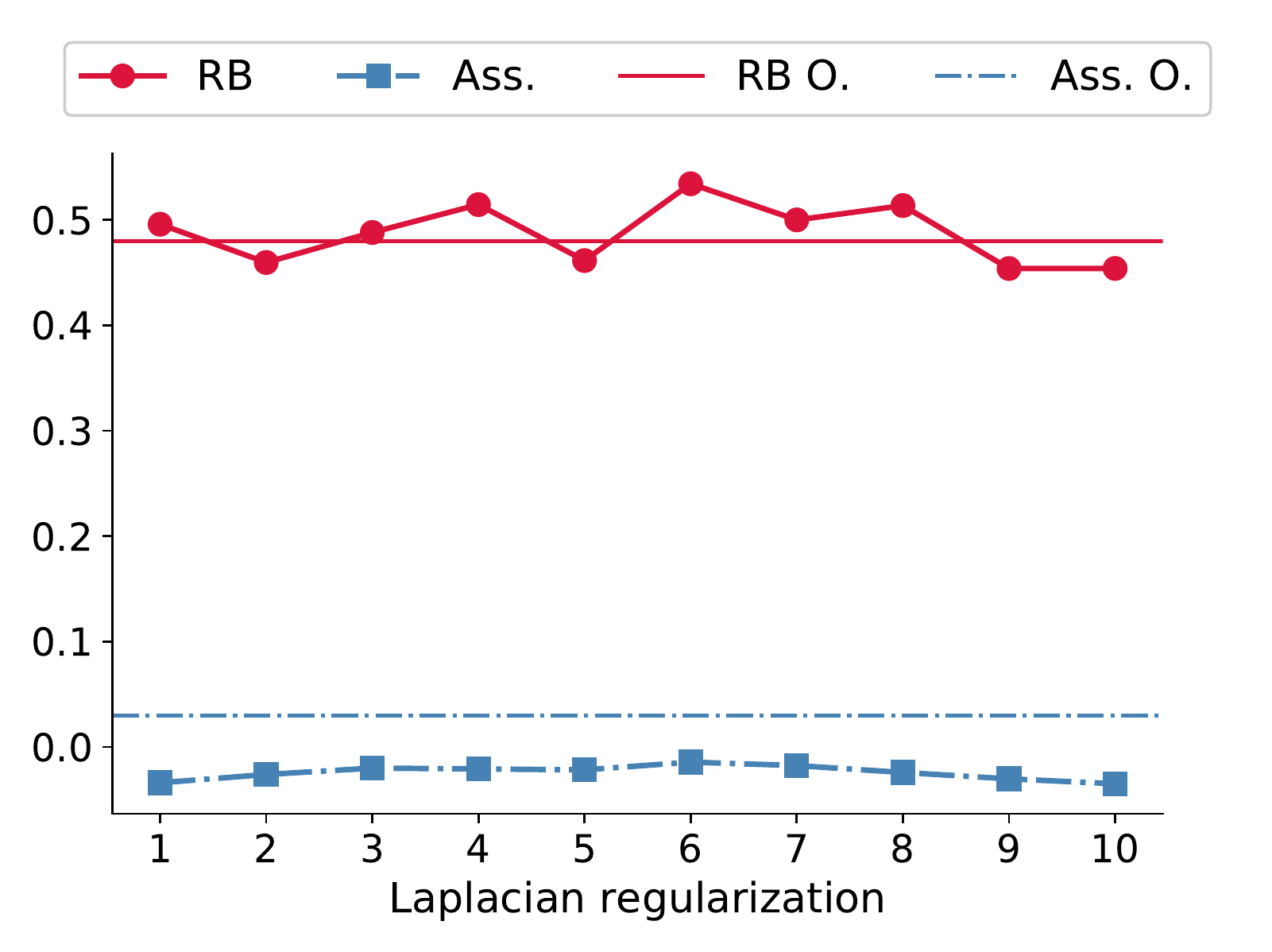}}
    \subfloat[\Gt]{\includegraphics[width=0.30\textwidth]{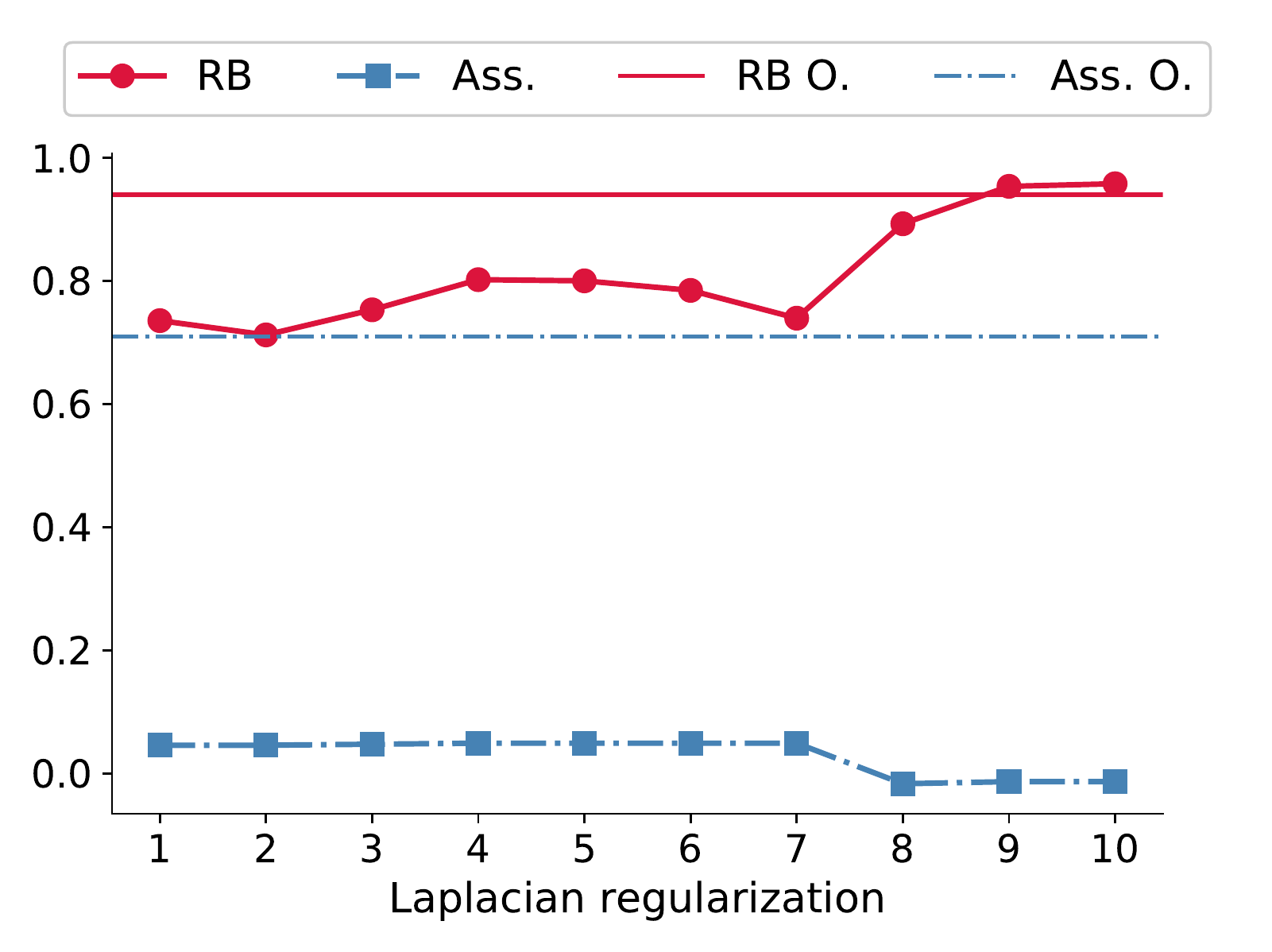}}\\
    \subfloat[\Gq]{\includegraphics[width=0.30\textwidth]{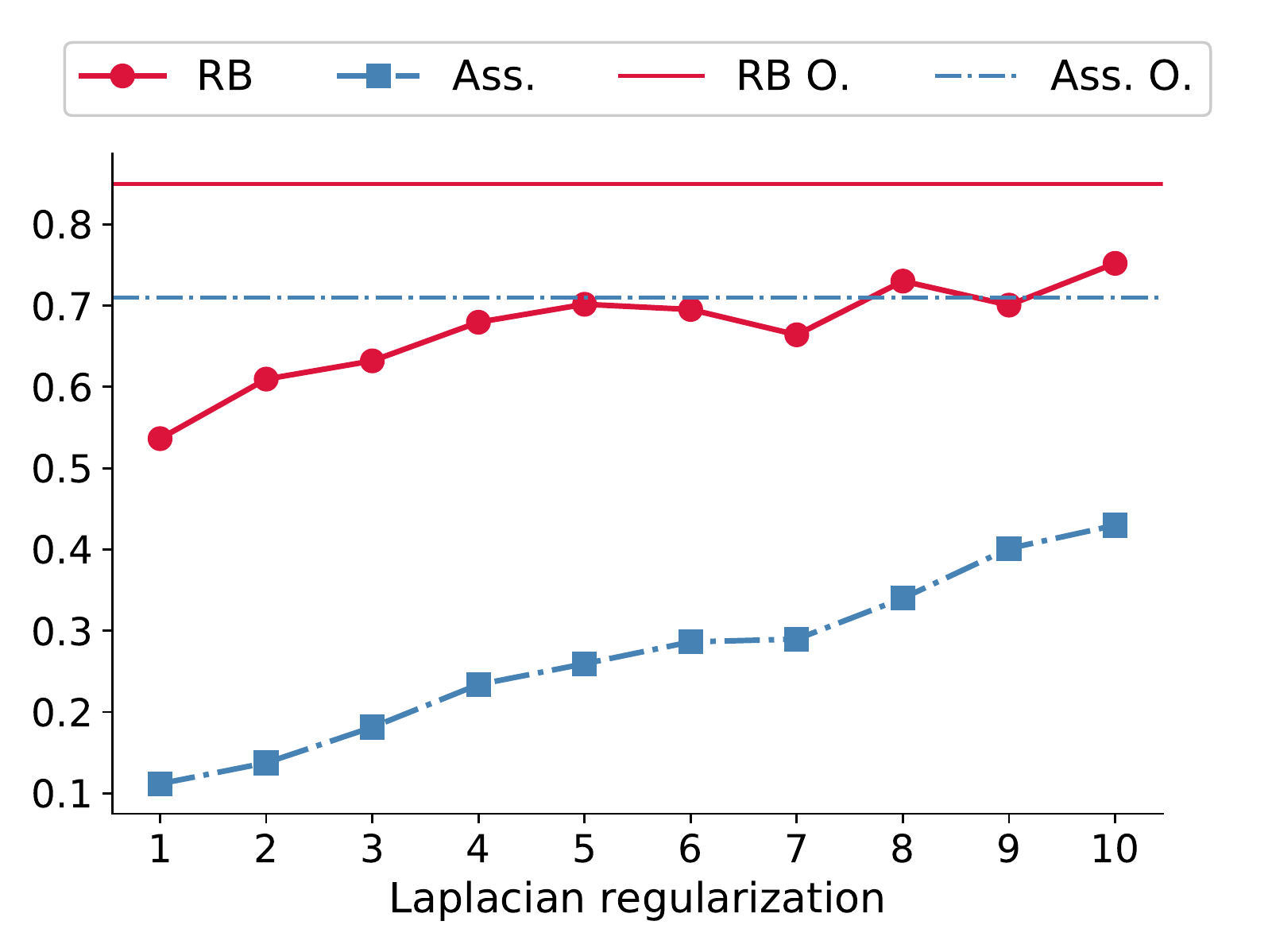}}
    \subfloat[\Gc]{\includegraphics[width=0.30\textwidth]{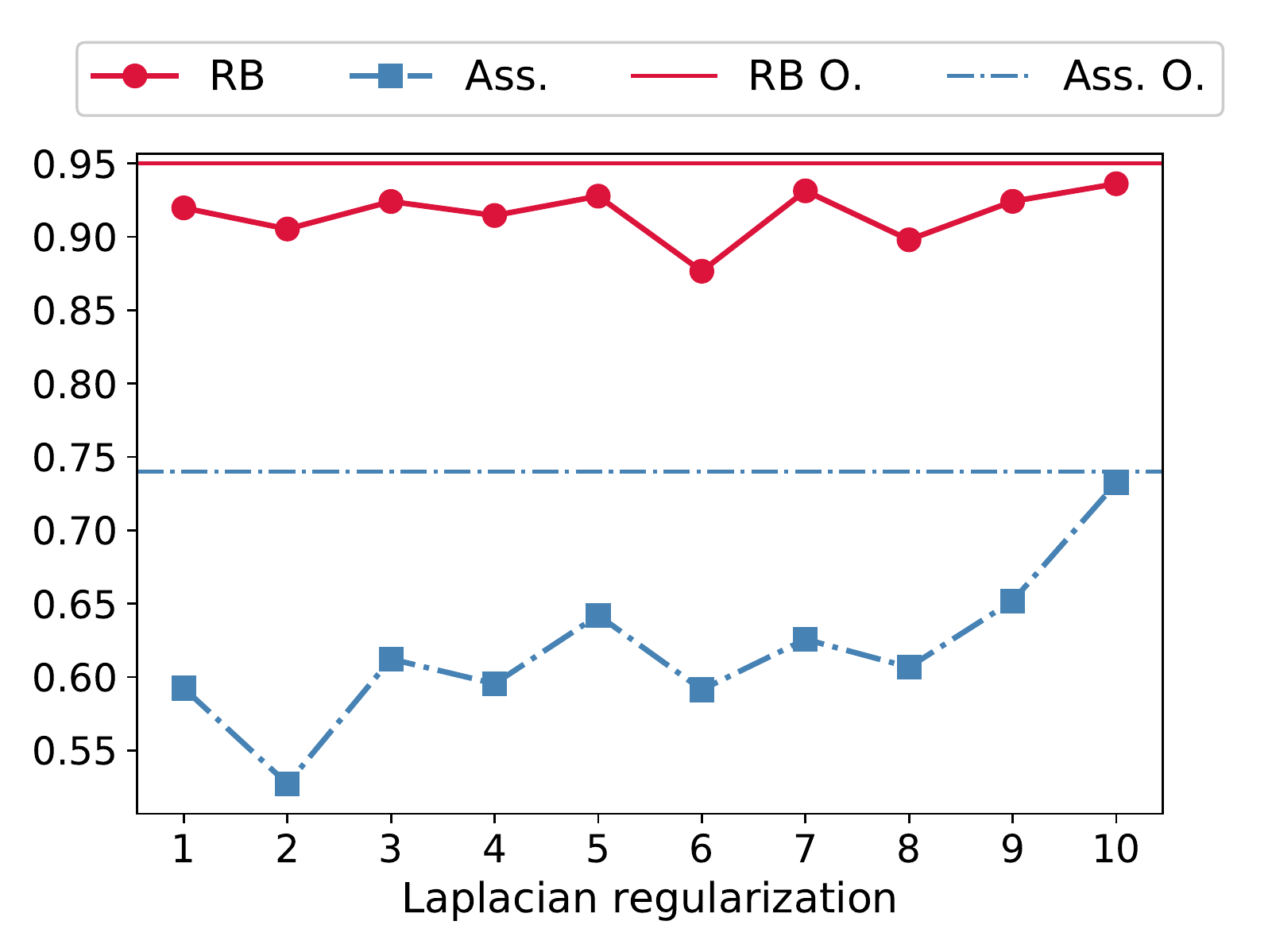}}
    \caption{Impact of the Laplacian regularisation on the RB and on the coefficient of assortativity (Ass.). The letter 'O' in the legend stands for original.}
    \label{fig:impact_reg}
\end{figure}

\paragraph{Role of the Laplacian regularization parameter}
As our algorithm depends on the regularization parameter $\lambda$ controlling the individual fairness of the repair, we proceed to its study on the synthetic graphs below. To this end, in Figure \ref{fig:impact_reg} we illustrates the effect of this parameter on the performance of our algorithm for the graphs considered above. Two major observations are in order here. First, for graphs {\Gu} and {\Gt} the strength of the Laplacian regularization leads to the increasing assortativity coefficient and RB scores below the original values, while for unbiased graph {\Gd} it has no particular effect on them. Second, the Laplacian regularization strength is less correlated with the quantities of interest in the imbalanced ({\Gt}) and multi-class settings {\Gc} which can be explained by the inherent drawbacks of optimal transport in this case. Dealing with these drawbacks presents an important open avenue for future research.

\subsection{Details on real-world networks}
Hereafter, we provide details for the real-world networks experiments reported in the main paper. 
\begin{itemize}
    \item For \textsc{N2Vec}-based approaches, we use the same set of default hyper-parameters: dimension of the embeddings is set to 64, the length of the walk to 15 and the window size is set to 10.
    For the link prediction, we train a logistic regression classifier on edge-wise feature vectors, where given a pair of users $(u, v)$ and their respective embeddings $z(u)$, $z(v)$, the goal is to predict the existence of an edge between $u$ and $v$. In what follows, edge-wise feature vectors correspond to the Hadamard product defined as element-wise multiplication between $z(u)$ and $z(v)$. We train the binary classifier by sampling non-existing edges as negative examples. 
    \item For \textsc{CNE}-based approaches, we use the set of default parameters recommended by the authors of \textsc{DeBayes} \cite{buyl2020}: dimension of the embeddings is set to 8 and the parameters $\sigma_1$ and $\sigma_2$ are set to 1 and 16, respectively. As for the link prediction, we follow their protocol by first computing the posterior $P(a_{ij}= 1|\text{train set})$ of the test links based on the embedding trained on the training network. Then the AUC score is computed by comparing the posterior probability of the test links and their true labels.
\end{itemize}

\bibliographystyle{unsrt}
\bibliography{_bib}

\begin{thebibliography}{10}

\bibitem{Feldman2015}
Michael Feldman, Sorelle~A. Friedler, John Moeller, Carlos Scheidegger, and
  Suresh Venkatasubramanian.
\newblock Certifying and removing disparate impact.
\newblock In {\em SIGKDD}, page 259–268, 2015.

\bibitem{Calmon2017}
Flavio Calmon, Dennis Wei, Bhanukiran Vinzamuri, Karthikeyan
  Natesan~Ramamurthy, and Kush~R Varshney.
\newblock Optimized pre-processing for discrimination prevention.
\newblock In {\em NeurIPS}, pages 3992--4001. 2017.

\bibitem{johndrow2019}
James~E. Johndrow and Kristian Lum.
\newblock An algorithm for removing sensitive information: Application to
  race-independent recidivism prediction.
\newblock {\em The Annals of Applied Statistics}, 13(1):189--220, 03 2019.

\bibitem{Zemel2013}
Rich Zemel, Yu~Wu, Kevin Swersky, Toni Pitassi, and Cynthia Dwork.
\newblock Learning fair representations.
\newblock In {\em ICML}, pages 325--333, 2013.

\bibitem{Edwards2016}
Harrison Edwards and Amos~J. Storkey.
\newblock Censoring representations with an adversary.
\newblock In {\em ICLR}, 2016.

\bibitem{Louizos2016}
Christos Louizos, Kevin Swersky, Yujia Li, Max Welling, and Richard~S. Zemel.
\newblock The variational fair autoencoder.
\newblock In {\em ICLR}, 2016.

\bibitem{Madras18}
David Madras, Elliot Creager, Toniann Pitassi, and Richard Zemel.
\newblock Learning adversarially fair and transferable representations.
\newblock In {\em ICML}, pages 3384--3393, 2018.

\bibitem{Gordaliza19}
Paula Gordaliza, Eustasio del Barrio, Fabrice Gamboa, and Jean{-}Michel Loubes.
\newblock Obtaining fairness using optimal transport theory.
\newblock In {\em ICML}, pages 2357--2365, 2019.

\bibitem{zafar2017a}
Muhammad~Bilal Zafar, Isabel Valera, Manuel Gomez~Rodriguez, and Krishna~P.
  Gummadi.
\newblock Fairness beyond disparate treatment \& disparate impact: Learning
  classification without disparate mistreatment.
\newblock In {\em WWW}, page 1171–1180, 2017.

\bibitem{zafar2017b}
Muhammad~Bilal Zafar, Isabel Valera, Manuel~Gomez Rogriguez, and Krishna~P.
  Gummadi.
\newblock {Fairness Constraints: Mechanisms for Fair Classification}.
\newblock In {\em AISTATS}, pages 962--970, 2017.

\bibitem{Corbett2017}
Sam Corbett-Davies, Emma Pierson, Avi Feller, Sharad Goel, and Aziz Huq.
\newblock Algorithmic decision making and the cost of fairness.
\newblock In {\em KDD}, page 797–806, 2017.

\bibitem{agarwal2018a}
Alekh Agarwal, Alina Beygelzimer, Miroslav Dudik, John Langford, and Hanna
  Wallach.
\newblock A reductions approach to fair classification.
\newblock In {\em ICML}, pages 60--69, 2018.

\bibitem{Donini2018}
Michele Donini, Luca Oneto, Shai Ben-David, John~S Shawe-Taylor, and
  Massimiliano Pontil.
\newblock Empirical risk minimization under fairness constraints.
\newblock In {\em NeurIPS}, pages 2791--2801. 2018.

\bibitem{Hardt2016}
Moritz Hardt, Eric Price, and Nati Srebro.
\newblock Equality of opportunity in supervised learning.
\newblock In {\em NeurIPS}, pages 3315--3323. 2016.

\bibitem{Kusner2017}
Matt~J Kusner, Joshua Loftus, Chris Russell, and Ricardo Silva.
\newblock Counterfactual fairness.
\newblock In {\em NeurIPS}, pages 4066--4076. 2017.

\bibitem{Jiang2019}
Ray Jiang, Aldo Pacchiano, Tom Stepleton, Heinrich Jiang, and Silvia Chiappa.
\newblock Wasserstein fair classification.
\newblock In {\em UAI}, page 315, 2019.

\bibitem{Chiappa19}
Silvia Chiappa.
\newblock Path-specific counterfactual fairness.
\newblock In {\em {AAAI}}, pages 7801--7808, 2019.

\bibitem{ZehlikeHW20}
Meike Zehlike, Philipp Hacker, and Emil Wiedemann.
\newblock Matching code and law: achieving algorithmic fairness with optimal
  transport.
\newblock {\em Data Min. Knowl. Discov.}, 34(1):163--200, 2020.

\bibitem{feldman15}
Michael Feldman, Sorelle~A. Friedler, John Moeller, Carlos Scheidegger, and
  Suresh Venkatasubramanian.
\newblock Certifying and removing disparate impact.
\newblock In {\em SIGKDD}, page 259–268, 2015.

\bibitem{DworkHPRZ12}
Cynthia Dwork, Moritz Hardt, Toniann Pitassi, Omer Reingold, and Richard~S.
  Zemel.
\newblock Fairness through awareness.
\newblock In {\em ITCS}, pages 214--226, 2012.

\bibitem{ferradans}
Sira Ferradans, Nicolas Papadakis, Julien Rabin, Gabriel Peyr{\'e}, and
  Jean-Fran{\c{c}}ois Aujol.
\newblock Regularized discrete optimal transport.
\newblock In Arjan Kuijper, Kristian Bredies, Thomas Pock, and Horst Bischof,
  editors, {\em Scale Space and Variational Methods in Computer Vision}, pages
  428--439, 2013.

\bibitem{opac-b1129524}
C\'edric Villani.
\newblock {\em Optimal transport : old and new}.
\newblock Springer, Berlin, 2009.

\bibitem{Cuturi14}
Marco Cuturi and Arnaud Doucet.
\newblock Fast computation of wasserstein barycenters.
\newblock In {\em ICML}, pages 685--693, 2014.

\bibitem{rahman2019}
Tahleen Rahman, Bartlomiej Surma, Michael Backes, and Yang Zhang.
\newblock Fairwalk: {Towards} {Fair} {Graph} {Embedding}.
\newblock In {\em IJCAI}, pages 3289--3295, 2019.

\bibitem{GroverL16}
Aditya Grover and Jure Leskovec.
\newblock node2vec: Scalable feature learning for networks.
\newblock In {\em SIGKDD}, pages 855--864, 2016.

\bibitem{buyl2020}
Maarten Buyl and Tijl~De Bie.
\newblock Debayes: a bayesian method for debiasing network embeddings.
\newblock In {\em Proceedings ICML}, pages 2537--2546, 2020.

\bibitem{Kang2019CNE}
Bo~Kang, Jefrey Lijffijt, and Tijl~De Bie.
\newblock Conditional network embeddings.
\newblock In {\em 7th International Conference on Learning Representations,
  {ICLR}}, 2019.

\bibitem{bose19a}
Avishek Bose and William Hamilton.
\newblock Compositional fairness constraints for graph embeddings.
\newblock In {\em ICML}, pages 715--724, 2019.

\bibitem{CourtyFTR17}
Nicolas Courty, R{\'{e}}mi Flamary, Devis Tuia, and Alain Rakotomamonjy.
\newblock Optimal transport for domain adaptation.
\newblock {\em {IEEE} Trans. Pattern Anal. Mach. Intell.}, 39(9):1853--1865,
  2017.

\bibitem{Adamic2005}
Lada~A. Adamic and Natalie Glance.
\newblock The political blogosphere and the 2004 u.s. election: Divided they
  blog.
\newblock In {\em 3rd International Workshop on Link Discovery}, page 36–43,
  2005.

\bibitem{Leskovec2012}
Jure Leskovec and Julian~J. Mcauley.
\newblock Learning to discover social circles in ego networks.
\newblock In {\em NIPS}, pages 539--547. Curran Associates, Inc., 2012.

\bibitem{QiuDMLWT18}
Jiezhong Qiu, Yuxiao Dong, Hao Ma, Jian Li, Kuansan Wang, and Jie Tang.
\newblock Network embedding as matrix factorization: Unifying deepwalk, line,
  pte, and node2vec.
\newblock In {\em WSDM}, pages 459--467, 2018.

\end{thebibliography}

\end{document}